\documentclass[10pt,twocolumn,letterpaper]{article}

\usepackage{cvpr}      % To produce the REVIEW version
\usepackage[accsupp]{axessibility}
\usepackage{graphicx}
\usepackage{amsmath}
\usepackage{amssymb}
\usepackage{booktabs}
\usepackage{bm}
\usepackage{enumitem}
\usepackage{multirow}
\usepackage{pifont}

\newcommand{\cmark}{\ding{51}}%
\newcommand{\xmark}{\ding{55}}%

\usepackage[pagebackref,breaklinks,colorlinks]{hyperref}

\usepackage[capitalize]{cleveref}
\crefname{section}{Sec.}{Secs.}
\Crefname{section}{Section}{Sections}
\Crefname{table}{Table}{Tables}
\crefname{table}{Tab.}{Tabs.}

\def\cvprPaperID{1729} % *** Enter the CVPR Paper ID here
\def\confName{CVPR}
\def\confYear{2022}

\begin{document}

\title{Neural Rays for Occlusion-aware Image-based Rendering}

\author{Yuan Liu$^1$ \quad Sida Peng$^2$ \quad  Lingjie Liu$^3$ \quad Qianqian Wang$^4$ \quad Peng Wang$^1$ \\ Christian Theobalt$^3$ \quad Xiaowei Zhou$^2$ \quad Wenping Wang$^{5}$ \\[0.3em]
$^1$The University of Hong Kong \quad $^2$Zhejiang University \quad $^3$Max Planck Institute for Informatics\\ $^4$Cornell University \quad $^5$Texas A\&M University}

\maketitle

%%%%%%%%% ABSTRACT
\begin{abstract}

We present a new neural representation, called {\em Neural Ray} (NeuRay), for the novel view synthesis task. Recent works construct radiance fields from image features of input views to render novel view images, which enables the generalization to new scenes. However, due to occlusions, a 3D point may be invisible to some input views. On such a 3D point, these generalization methods will include inconsistent image features from invisible views, which interfere with the radiance field construction. To solve this problem, we predict the visibility of 3D points to input views within our NeuRay representation. This visibility enables the radiance field construction to focus on visible image features, which significantly improves its rendering quality. Meanwhile, a novel consistency loss is proposed to refine the visibility in NeuRay when finetuning on a specific scene. Experiments demonstrate that our approach achieves state-of-the-art performance on the novel view synthesis task when generalizing to unseen scenes and outperforms per-scene optimization methods after finetuning. Project page: \href{https://liuyuan-pal.github.io/NeuRay/}{https://liuyuan-pal.github.io/NeuRay/
\vspace{-10pt}}

% We present a new neural representation, called {\em Neural Ray} (NeuRay), for the novel view synthesis (NVS) task. 
% Recent works construct radiance fields from image features of input views for the NVS task, which enables the generalization to new scenes but has difficulty in handling occlusions. 
% NeuRay depicts scenes by probability distributions defined on rays, which enables efficient computation of visibility for occlusion inference to improve the NVS quality. 
% In generalization to unseen scene, the distribution parameters in NeuRay can be effectively estimated from a cost volume construction.
% In finetuning NeuRay on a specific scene, a novel consistency loss is proposed to refine the NeuRay representation better and better.
% Moreover, probability distributions in NeuRay also enable a fast estimation of surface locations, which can be used in speeding up the rendering process.
% To address this issue, NeuRay serves as a backend representation for these methods and enables consideration of visibility for occlusion inference. 
% When generalize to unseen scenes, 
% This distribution enables us to efficiently infer the visibility from input views to any 3D point with one forward pass of the network and the distribution parameters can be effectively estimated from a cost volume construction. 
% Experiments demonstrate that our approach achieves state-of-the-art performance on the NVS task when generalizing to unseen scenes and outperforms per-scene optimization methods after finetuning on each scene.

\end{abstract}

\section{Introduction}

Novel View Synthesis (NVS) is an important problem in computer graphics and computer vision. Given a set of input images with known camera poses, the goal of NVS is to synthesize images of the scene from arbitrary virtual camera poses. Recently, neural rendering methods have achieved impressive improvements on the NVS problem compared to earlier image-based rendering methods~\cite{kalantari2016learning, hedman2018deep, mildenhall2019local}. Neural Radiance Field (NeRF)~\cite{mildenhall2020nerf} shows that photo-realistic images of novel views can be synthesized by volume rendering on a 5D \textit{radiance field} encoded in a neural network which maps a position and a direction to a density and a color. However, these methods cannot generalize to unseen scenes as they learn scene-specific networks, which usually take hours or days for a single scene.

\begin{figure}
    \centering
    \includegraphics[width=\linewidth]{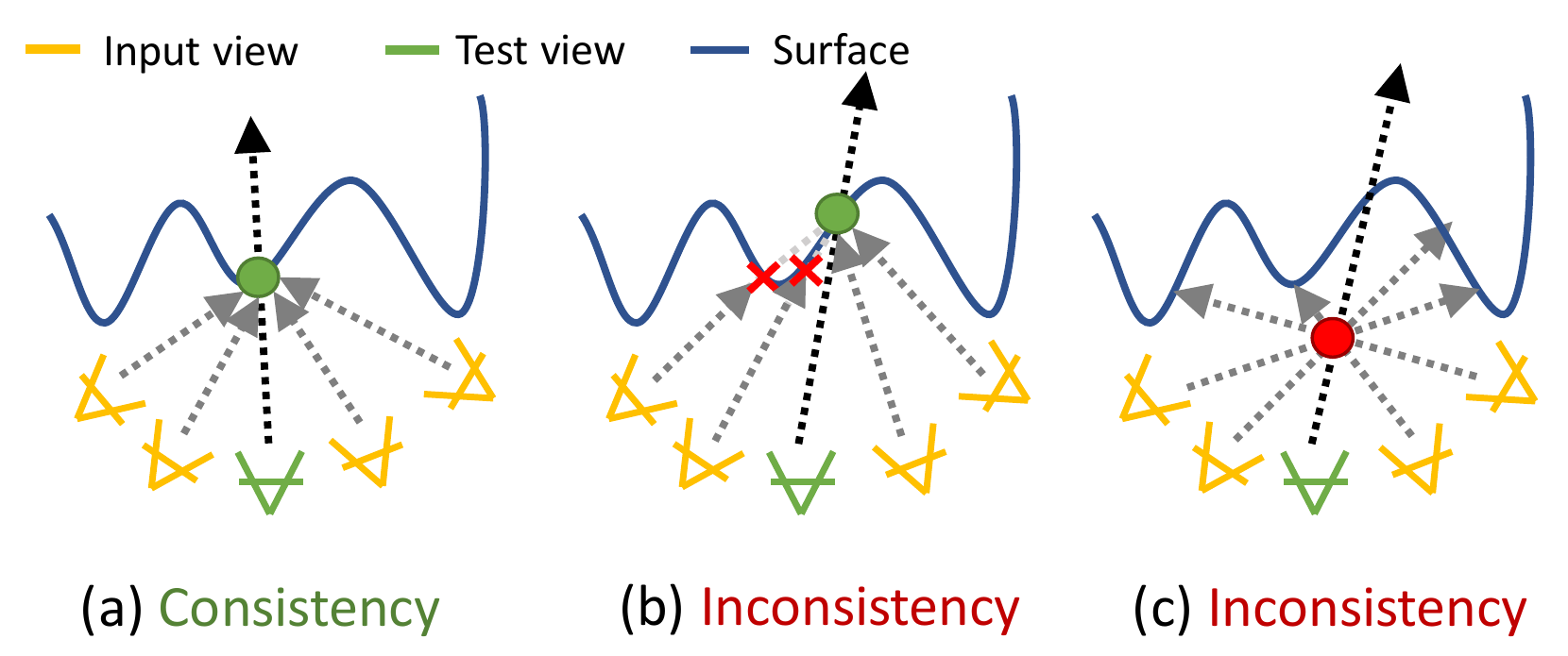}
    \caption{(a) Without occlusions, local image features are consistent on the surface point. (b) Local image features are inconsistent on a surface point due to occlusions. (c) Local image features are inconsistent on a non-surface point. 
    Generalization methods will correctly assign a large density to the surface point in (a) due to the feature consistency. However, when features are not very consistent in (b) and (c), it is relatively hard for these methods to correctly determine the density.}
    \label{fig:occ}
    \vspace{-15pt}
\end{figure}

Recent works \cite{Trevithick20arxiv_GRF, wang2021ibrnet, chibane2021stereo, Yu20arxiv_pixelNeRF} propose NeRF-like neural rendering frameworks that can generalize to unseen scenes. Given a set of input views, they construct a radiance field on-the-fly by extracting local image features on these views and matching multi-view features to predict colors and density on 3D points. This is similar to traditional stereo matching methods \cite{schoenberger2016mvs, yao2018mvsnet} that check the multi-view feature consistency to find a surface point, as shown in Fig.~\ref{fig:occ} (a). However, when features are not consistent on a point, it is relatively hard for these methods to correctly determine whether such inconsistency is caused by occlusions as shown in Fig.~\ref{fig:occ} (b) or this point is a non-surface point as shown in Fig.~\ref{fig:occ} (c), leading to rendering artifacts.

To address this problem, we introduce a new neural representation called Neural Rays (NeuRay) in this paper. 
NeuRay consists of pixel-aligned feature vectors on every input view. On a camera ray emitting from a pixel on the input view, the associated NeuRay feature vector on this pixel is able to predict visibility to determine whether a 3D point at a specific depth is visible or not. With such visibility, we can easily distinguish the occlusion-caused feature inconsistency from the non-surface-caused feature inconsistency in Fig.~\ref{fig:occ}, which leads to more accurate radiance field construction and thus better rendering quality on difficult scenes with severe self-occlusions.

A key challenge is how to estimate the visibility in an unseen scene. This is a chicken-and-egg problem because the estimation of visibility requires knowing the surface locations while the estimated visibility is intended for better surface estimation in the radiance field construction. To break this cycle, we propose to apply well-engineered multi-view stereo (MVS) algorithms, like cost volume construction~\cite{yao2018mvsnet} or patch-matching~\cite{schoenberger2016mvs}, to reconstruct the scene geometry and then extract the pixel-aligned feature vectors of NeuRay from the reconstructed geometry. In the end, NeuRay will be used in the computation of the visibility to improve the radiance field construction.

Another problem is how to parameterize such visibility in NeuRay. A direct way is to predict the densities along the camera ray from the view to the 3D point and then accumulate these densities to compute the transmittance as the visibility like NeRF~\cite{mildenhall2020nerf}. However, computing visibility with this strategy is computationally impractical because given $N$ input views and a 3D point, we should accumulate the density along all $N$ camera rays from every input view to this 3D point, which means we need to sample $K$ points on every camera ray and evaluate the density on all $N\times K$ sample points. To reduce the computation complexity, we directly parameterize the visibility with a Cumulative Distribution Function (CDF) in NeuRay, which avoids density accumulation along rays and only requires $N$ network forward passes to compute the visibility of all $N$ input views.

NeuRay not only help the radiance field construction on unseen scenes but also be able to refine itself by finetuning on a specific scene with a novel consistency loss. Since both the NeuRay representation and the constructed radiance field depict the scene geometry, we propose a loss to enforce the consistency between the surface locations from the NeuRay representation and those from the constructed radiance field. This loss enables NeuRay to memorize the scene geometry predicted by the radiance field. At the same time, the memorized scene geometry in NeuRay will in turn improve the radiance field construction by providing better occlusion inference.
% Based on the probability distribution defined on the camera ray, NeuRay is able to estimate the surface location on the ray.
% Meanwhile, the constructed radiance field is also able to compute surface locations on camera rays.
% % Meanwhile, constructing radiance fields also depicts the scene geometry by predicting surface locations. 
% Thus, in the scene-specific optimization of NeuRay, we propose a novel consistency loss to enforce the consistency between the surface locations from the NeuRay representation and those from the constructed radiance field. This loss enables the NeuRay representation to refine itself by memorizing the scene geometry predicted by the radiance field. At the same time, the memorized scene geometry in NeuRay will in turn improves the radiance field construction by providing better occlusion inference.
% Another advantage is that estimating surface locations by NeuRay is efficient and can be used in speeding up the rendering process. Using the estimated surface locations, we are able to quickly filter out non-surface points and only apply multi-view feature matching on few remaining points for fast rendering.

We conducted extensive experiments on the NeRF synthetic dataset~\cite{mildenhall2020nerf}, the DTU dataset~\cite{jensen2014large} and the LLFF dataset~\cite{mildenhall2019local} to demonstrate the effectiveness of NeuRay. The results show that 1) without scene-specific optimization, our method already produces satisfactory rendering results that outperforms other generalization methods by a large margin; 2) finetuning NeuRay produces much superior results than finetuning other generalization models and achieve even better rendering quality than NeRF~\cite{mildenhall2020nerf}.
Moreover, we can speed up rendering with the help of NeuRay by caching features on input views and predicting coarse surface locations, which costs $\sim$3 seconds to render an image of size $800\times600$.

\section{Related works}

\subsection{Image-based rendering}

Many works \cite{gortler1996lumigraph, levoy1996light, chaurasia2013depth, kalantari2016learning, hedman2016scalable, hedman2018deep, Riegler2020FVS, kopanas2021point} have focused on blending input images with geometry proxies to synthesize novel views. Conventional light field-based methods \cite{gortler1996lumigraph, davis2012unstructured} reconstruct a 4D plenoptic function from densely sampled views, which achieve photo-realistic rendering results but typically have a limited renderable range. To extend the renderable range, some works \cite{chaurasia2013depth, penner2017soft} seek the help of 3D proxy geometry from multi-view stereo (MVS) methods \cite{schoenberger2016mvs}. With the development of deep learning techniques, some methods \cite{kalantari2016learning, hedman2018deep, choi2019extreme, thies2020ignor, xu2019deep, Riegler2020FVS, Riegler2021SVS} introduce convolutional neural networks (CNNs) to replace hand-crafted components of the image-based rendering (IBR). One common challenge for the IBR methods is the sensitivity to the quality of estimated depth maps~\cite{choi2019extreme}. Our method also belongs to the category of image-based rendering and also uses cost volumes or estimated depth from MVS methods. However, our method can be trained from scratch without the help of external MVS algorithms and also can be finetuned on a scene to remedy reconstruction errors of MVS. 
% Concurrent works~\cite{ruckert2021adop, kopanas2021point} represent scene by point clouds, which can also be initialized from external MVS methods and further optimized by differentiable rendering.

\subsection{Neural scene representation}
Recently, instead of estimating an external 3D proxy geometries, some methods have attempted to construct explicit trainable 3D representations from input images with differentiable renderers, such as voxels \cite{sitzmann2019deepvoxels, lombardi2019neural}, textured meshes \cite{thies2019deferred, liu2019neural, liu2020NeuralHumanRendering,habermann2021}, and point clouds \cite{wu2020multi, aliev2020neural, ruckert2021adop, kopanas2021point}. Then, novel view can be synthesized from the constructed 3D representations. To further improve the rendering resolution, some methods \cite{sitzmann2019scene, mildenhall2020nerf, liu2020nsvf, peng2021neural, liu2020dist, niemeyer2020differentiable,Kellnhofer2021nlr} resort to pure neural fields encoded by neural networks to represent 3D scenes. Our method represents a scene by a ray-based representation for the NVS task. PIFu~\cite{saito2019pifu} and its follow-ups~\cite{saito2020pifuhd,hong2021stereopifu,he2020geo} also use ray-based representations to reconstruct human shapes. 
NeRF \cite{mildenhall2020nerf} renders photo-realistic images by volume rendering on a radiance field. Many following works \cite{liu2020nsvf, garbin2021fastnerf, reiser2021kilonerf, yu2021plenoctrees, MartinBrualla20arxiv_nerfw, Niemeyer20arxiv_GIRAFFE, peng2021animatable,liu2021neuralactor,Wizadwongsa2021NeX,srinivasan2021nerv} have attempted to improve NeRF in various aspects. Among these, NeRV~\cite{srinivasan2021nerv} also uses a visibility prediction for efficient relighting. In comparison, visibility in our method is used in efficiently constructing a radiance field on-the-fly.

\textbf{Generalization volume rendering}. NeRF usually takes a long time to train on each new scene. To address this, recent works \cite{Trevithick20arxiv_GRF, Yu20arxiv_pixelNeRF, raj2021pva, Rematas21arxiv_sharf, wang2021ibrnet, chibane2021stereo, chen2021mvsnerf} introduce generalizable rendering methods which construct a radiance field on-the-fly. Also, some works speed up NeRF-training by meta-learning~\cite{bergman2021metanlr,tancik2021learned} or voxel~\cite{muller2022instant,sun2021direct,yu2021plenoxels}. 
Our method also bases on constructing radiance field on-the-fly to generalize to unseen scenes. The difference is that our method uses a ray-based representation for occlusion inference which greatly improves the rendering quality.
% NeuRay also adopts the volume rendering approach from NeRF and uses stereo matching in the pipeline, but NeuRay can be initialized from cost volumes, which saves time for scene-specific training. As a backend representation of stereo matching based methods, NeuRay encodes visibilities for the occlusion inference to improve the rendering quality.

\section{Method}

Given $N$ input views of a scene with known camera poses, our goal is to render images on arbitrary novel test views. Before introducing NeuRay, we first review the volume rendering on a radiance field~\cite{mildenhall2020nerf}. 

\begin{figure}
    \centering
    \includegraphics[width=\linewidth]{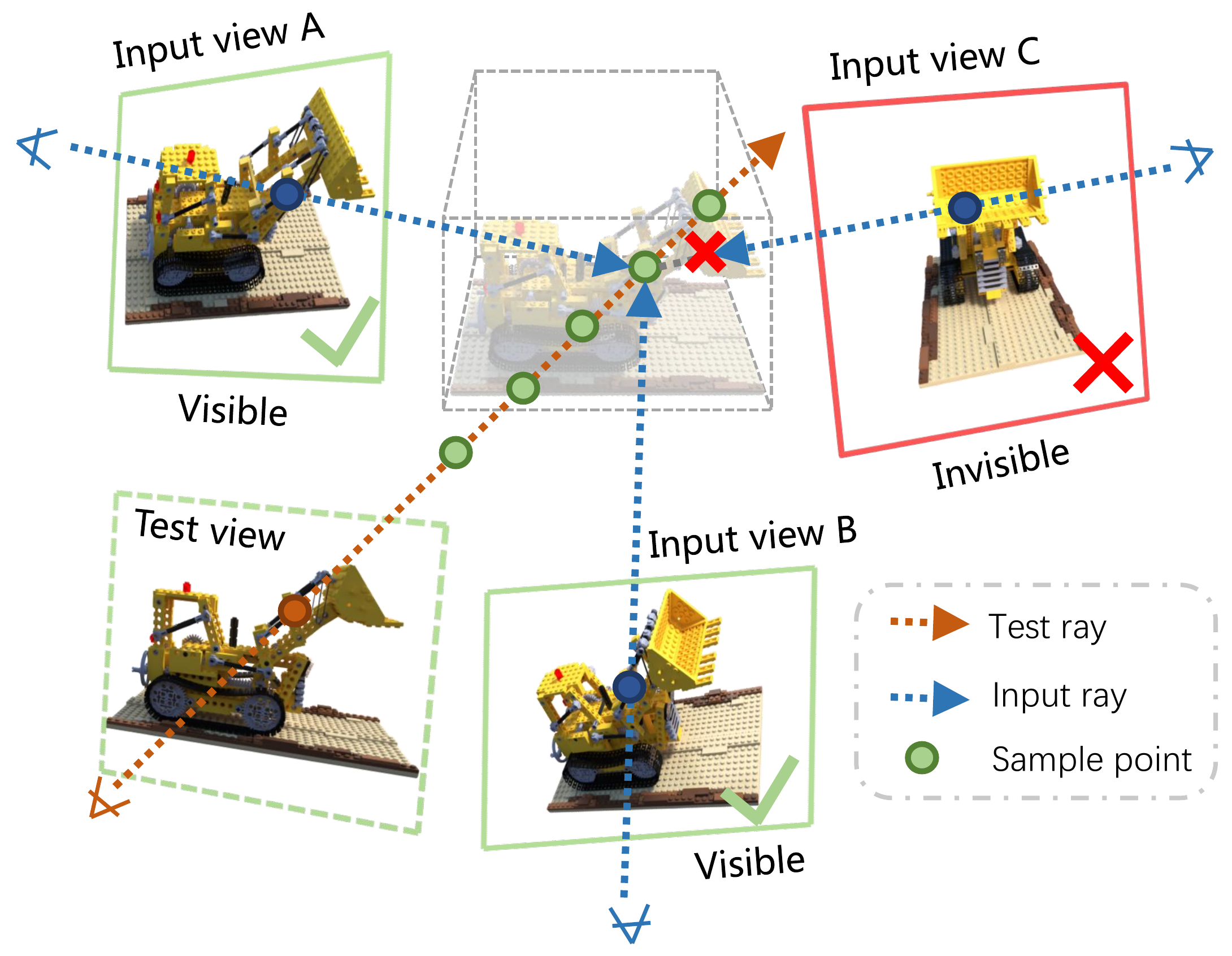}
    \caption{Constructing a radiance field on-the-fly from input views to synthesize the test view image by volume rendering. We first sample points on the test ray, then aggregate local features of input views to determine the alpha values and colors of sample points, and finally accumulating colors by volume rendering to compute the output color. 
    Our method constructs NeuRay on input views to predict the visibility to sample points so that it allows occlusion-aware feature aggregation on these sample points.}
    \label{fig:construct}
    \vspace{-10pt}
\end{figure}

\subsection{Volume rendering}
\label{sec:vr}
In our method, images of test views are synthesized by volume rendering as shown in Fig.~\ref{fig:construct}.
Assuming a camera ray emitting from the test view, called a \textit{test ray}, is parameterized by $\bm{p}(z) =\bm{o}+z\bm{r}$, $z\in \mathbb{R}^{+}$, where $\bm{o}$ is the start point at the camera center, and $\bm{r}$ is the unit direction vector of the ray. First, we sample $K_t$ points $\{\bm{p}_i \equiv \bm{p}(z_i) |i=1,...,K_t\}$ with the increasing values $z_i$ along the ray. Then, the color for the associated pixel of this camera ray is computed by
\begin{equation}
    \bm{c}=\sum_{i=1}^{K_t} \bm{c}_{i} h_{i},
    \label{eq:vr}
\end{equation}
where $\bm{c} \in \mathbb{R}^3$ is the rendered color for the pixel, $\bm{c}_{i}\in \mathbb{R}^3$ is the color of the sample point $\bm{p}_i$, and $h_{i}\in \mathbb{R}$ the hitting probability that the ray is not occluded by any depth up to the depth $z_i$ and hits a surface in the range $(z_i,z_{i+1})$. Thus, the hitting probability $h_{i}$ can be computed by
\begin{equation}
    h_{i}=\prod_{k=1}^{i-1} (1-\alpha_k)\alpha_i,
    \label{eq:prod}
\end{equation}
where $\alpha_i$ is the alpha value in the depth range $(z_i, z_{i+1})$. 
In order to render a novel image using Eq.~(\ref{eq:vr}) and Eq.~(\ref{eq:prod}), we construct a radiance field to compute $\alpha_i$ and $\bm{c}_i$.

\subsection{Occlusion-aware radiance field construction}
\label{sec:construct}
\textbf{Radiance field construction on-the-fly}. In contrast to NeRF~\cite{mildenhall2020nerf} which learns a scene-specific neural radiance field, generalization rendering methods~\cite{wang2021ibrnet,Yu20arxiv_pixelNeRF,chibane2021stereo,Trevithick20arxiv_GRF} construct a radiance field on-the-fly by aggregating local features. Given a 3D point $\bm{p}_i\in \mathbb{R}^3$ as shown in Fig.~\ref{fig:construct}, these methods first extract features on input views by a CNN and then aggregate features of input views on this point by
\begin{equation}
    \bm{f}_i = \mathcal{M} (\{\bm{f}_{i,j}|j=1,...,N\}),
    \label{eq:agg}
\end{equation}
where $\bm{f}_{i,j}$ is the local image feature of the sample point $\bm{p}_i$ projected on the $j$-th input view, $\mathcal{M}$ is a network which aggregates the features from different views to produce a feature $\bm{f}_i$. Then, the alpha value $\alpha_i$ and color $\bm{c}_i$ for this point $\bm{p}_i$ will be decoded from the aggregated feature $\bm{f}_i$ by other networks~\cite{wang2021ibrnet,Yu20arxiv_pixelNeRF,chibane2021stereo,Trevithick20arxiv_GRF}. We provide more details about this in the appendix.

\textbf{Occlusion-aware construction}. The proposed method also constructs a radiance field on-the-fly as previous methods. Additionally, we predict a visibility term $v_{i,j}$ illustrating the $j$-th input view is visible or not to this 3D point $\bm{p}_i$ for the occlusion-aware feature aggregation
\begin{equation}
    \bm{f}_i = \mathcal{M} (\{\bm{f}_{i,j},v_{i,j}|j=1,...,N\}).
    \label{eq:aggvis}
\end{equation}
In this case, the aggregation network $\mathcal{M}$ is able to focus on visible views in the aggregated feature $\bm{f}_i$ and reduces the interference from invisible views. In the following, we will introduce our NeuRay for the computation of visibility $v_{i,j}$.

\subsection{NeuRay representation}
\label{sec:neuray}
\begin{figure}
    \centering
    \includegraphics[width=\linewidth]{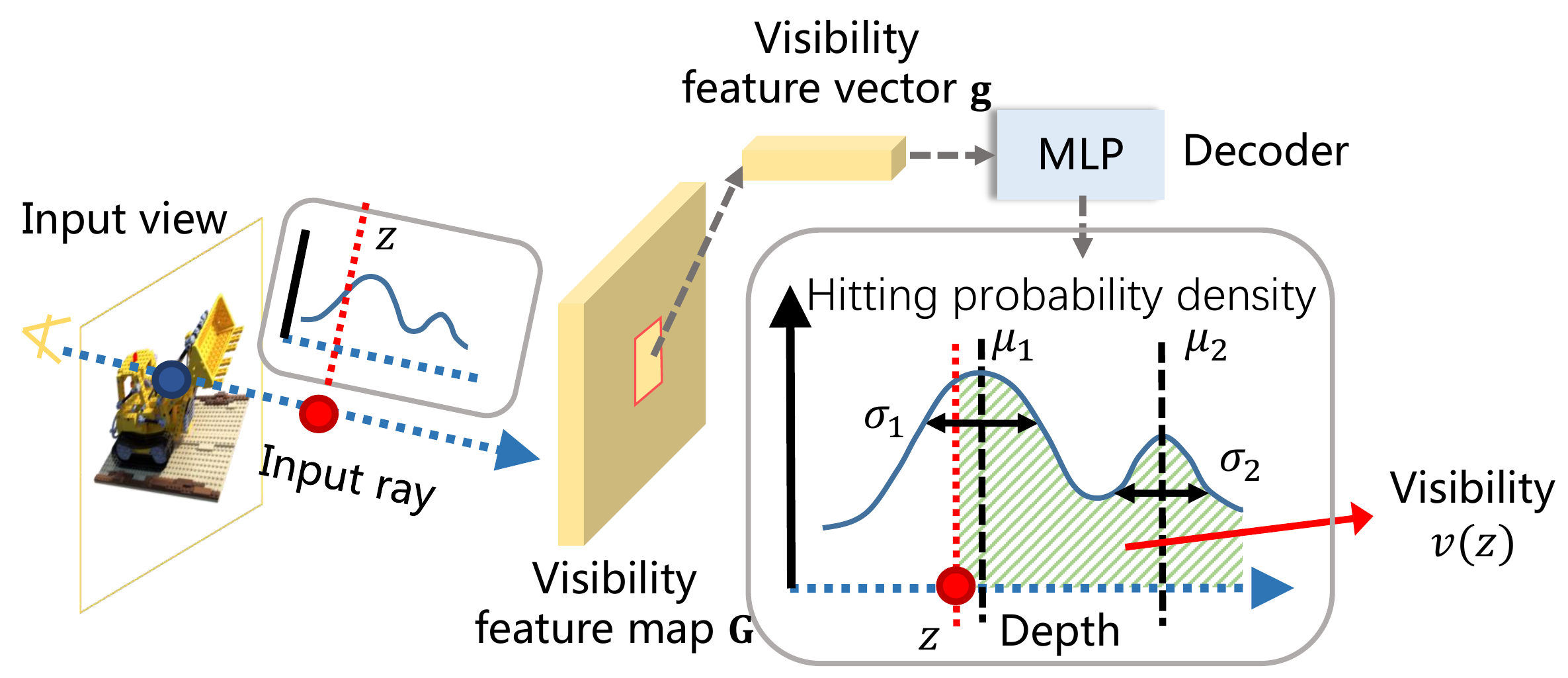}
    \caption{Visibility computation in NeuRay. NeuRay consists of a visibility feature map $\bm{G}$ on every input view. Every feature vector $\bm{g}$ on $\bm{G}$ can be decoded to a mixture of logistics distributions by an MLP. The distribution illustrates the visibility function $v(z)$ of the input ray emitting from the location of $\bm{g}$, which is the area under the curve after the depth $z$.}
    \label{fig:vis}
    \vspace{-10pt}
\end{figure}

Given a camera ray emitting from a input view, called an \textit{input ray}, NeuRay is able to predict the visibility function $v(z)$ indicating a point at depth $z$ is visible or not for this input ray as shown in Fig.~\ref{fig:vis}. 
On every input view, NeuRay is represented by a visibility feature map $\bm{G}\in \mathbb{R}^{H\times W\times C}$. 
Denoting $\bm{g}\in \mathbb{R}^C$ as a corresponding feature vector for the given input ray on $\bm{G}$, we will compute the visibility $v(z)$ of this input ray from $\bm{g}$.
Obviously, a valid visibility function $v(z)$ should be non-increasing on $z$ and $0\le v(z)\le 1$. In the following, we discuss how to parameterize the visibility function $v(z)$ from $\bm{g}$. Then, we will introduce how to compute $\bm{G}$ in Sec.~\ref{sec:gen} and Sec.~\ref{sec:ft}.

\begin{figure*}
    \centering
    \includegraphics[width=\textwidth]{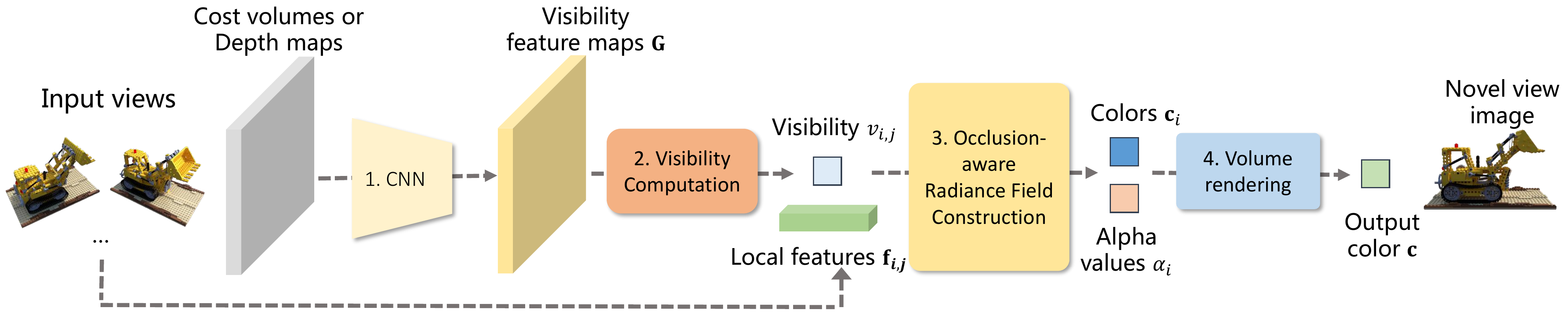}
    \caption{Pipeline of rendering with NeuRay. 1. On input views, cost volumes or depth maps are estimated, which are used in predicting visibility feature maps by a CNN (Sec.~\ref{sec:gen}). 2. Visibility feature maps are used in the computation of the visibility of input views to 3D points (Sec.~\ref{sec:neuray}). 3. For 3D points, we aggregate local features from input views along with the visibility to compute alpha values and colors on these points (Sec.~\ref{sec:construct}). 4. Volume rendering is applied to accumulate alpha values and colors to synthesize images (Sec.~\ref{sec:vr}). }
    \label{fig:gen_pipeline}
    \vspace{-10pt}
\end{figure*}

\textbf{Visibility from occlusion probability}. We represent the visibility function with a Cumulative Density Function (CDF) $t(z)$ by $v(z)=1-t(z)$ as shown in Fig.~\ref{fig:vis}, which is parameterized as a mixture of logistics distributions
\begin{equation}
    t(z;\{\mu_i,\sigma_i,w_i\})=\sum_i^{N_{l}} w_i S((z-\mu_i)/\sigma_i),
    \label{eq:occlusion}
\end{equation}
where we mix $N_{l}$ logistics distributions, $\mu_i$ is the mean of $i$-th logistics distribution, $\sigma_i$ the standard deviation, $w_i$ the mixing weight with $\sum_i w_i=1$, $S(\cdot)$ a sigmoid function. All parameters $[\mu_i,\sigma_i,w_i]=\mathcal{F}(\bm{g})$ are decoded from the feature $\bm{g}$ by an MLP $\mathcal{F}$. As a CDF, $t(z)$ is non-decreasing, so that $1-t(z)$ forms a valid visibility function. 

$t(z)$ actually corresponds to an \textit{occlusion probability} that the input ray is occluded before a depth $z$ and we call the corresponding probability density function (PDF) \textit{hitting probability density}. The visibility $v(z)$ is actually the area under the PDF curve after $z$.
In general, a ray will only hit one surface so one logistics distribution will be enough. However, using a mixture of logistics distributions improves performance when the ray hits on semi-transparent surfaces or edges of surfaces.

\textbf{Discussion}. Alternatively, we may parameterize visibility with a NeRF~\cite{mildenhall2020nerf}-like density. However, computing visibility with this strategy is too computationally intensive. In this formulation, we directly decode a density $d(z)=\phi(z;\bm{g})$ from $\bm{g}$ using a MLP $\phi$. To compute the visibility $v(z)$, we need to first sample $K_r$ depth $\{z_k\}$ with $z_k<z$, compute their densities $d(z_k)$ and the corresponding alpha values $\alpha_{k}=1-\exp(-{\rm ReLU}(d_k))$, and finally get visibility $v(z)=\prod_{k=1}^{K_r} (1-\alpha_k)$. Though the formulation is a valid visibility function, it is computationally impractical because it requires $K_r$ times evaluation of $\phi$ to compute the visibility of a input view to a point.

\subsection{Generalize with NeuRay}
\label{sec:gen}

When rendering in unseen scenes, we extract the visibility feature maps $\bm{G}$ from a cost volume construction~\cite{yao2018mvsnet} or a patch-match stereo~\cite{schoenberger2016mvs}. 
In the cost volume construction on every input view, we use its $N_s$ neighboring input views to construct a cost volume of size $H\times W \times D$~\cite{yao2018mvsnet}. 
Then, a CNN is applied on the cost volume to produce the visibility feature map $\bm{G}\in \mathbb{R}^{H\times W\times C}$ for this input view. 
Alternatively, we can also directly extract feature maps $\bm{G}$ from depth maps estimated by patch-match stereo~\cite{schoenberger2016mvs}, in which the estimated depth map of size ${H\times W}$ is processed by a CNN to produce the visibility feature map $\bm{G}$.

\textbf{Pipeline}. The whole pipeline of rendering with NeuRay in an unseen scene is shown in Fig.~\ref{fig:gen_pipeline}. On all input views, cost volumes or depth maps are estimated by MVS algorithms~\cite{yao2018mvsnet,schoenberger2016mvs}, which are processed by a CNN to produce visibility feature maps $\bm{G}$. Then, for 3D sample points on test rays, we compute the visibility $v_{i,j}$ of input views to these points (Sec.~\ref{sec:neuray}) and aggregate the local features $\bm{f}_{i,j}$ along with $v_{i,j}$ to compute the alpha values and colors on these points (Sec.~\ref{sec:construct}). Finally, the alpha values and colors are accumulated along the test rays by volume rendering to synthesize the test images (Sec.~\ref{sec:vr}).

\textbf{Loss}. The whole rendering framework can be pretrained on training scenes and then directly applied on unseen scenes for rendering. To pretrain the rendering framework, we randomly select a view in a training scene as the test view and use other views as input views to render the selected test view with a render loss
\begin{equation}
    \ell_{render}=\sum \|\bm{c} - \bm{c}_{gt}\|^2,
    \label{eq:render_loss}
\end{equation}
where $\bm{c}$ is computed by Eq.~\ref{eq:vr}, $\bm{c}_{gt}$ is the ground-truth color. 

\begin{figure}
    \centering
    \includegraphics[width=\linewidth]{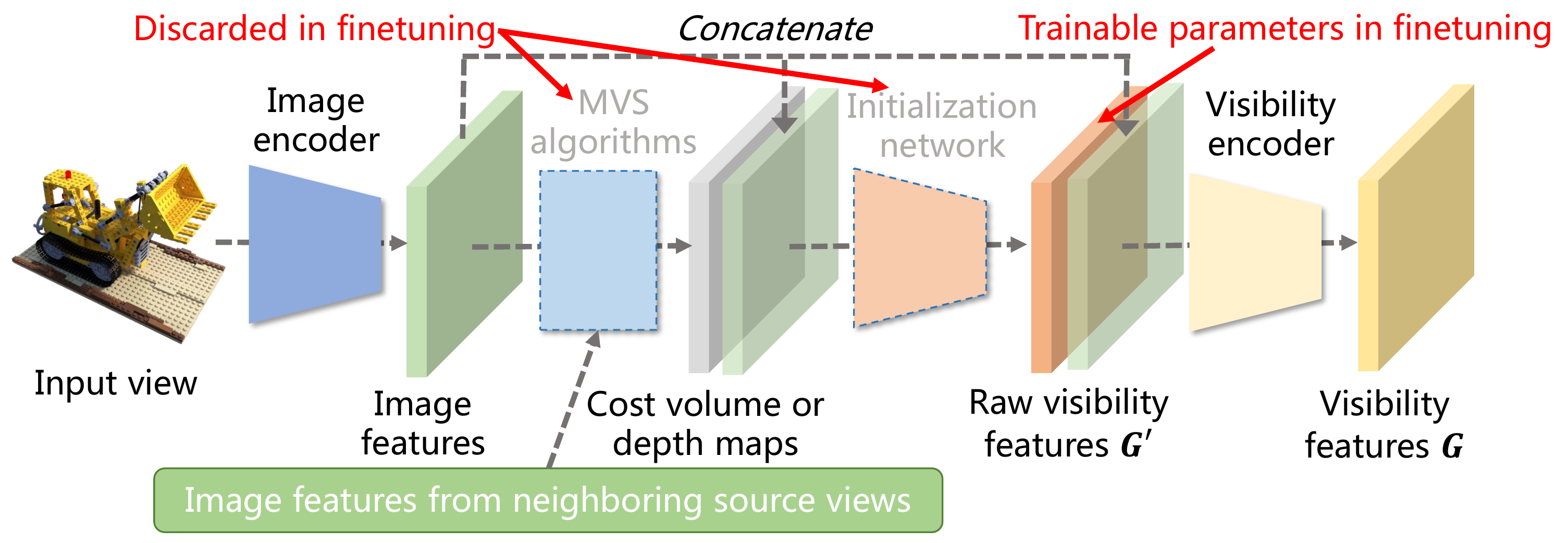}
    \caption{When finetuning on a scene, we treat the intermediate feature map $\bm{G}'$ as trainable parameters of NeuRay, which is initialized by the initialization network. Then, the initialization network will be discarded in finetuning.}
    \label{fig:ft}
    \vspace{-10pt}
\end{figure}

\subsection{Finetune with NeuRay}
\label{sec:ft}

As done in previous works~\cite{wang2021ibrnet,Yu20arxiv_pixelNeRF,chibane2021stereo,Trevithick20arxiv_GRF}, the proposed rendering framework can be further finetuned on a specific scene to achieve better rendering quality on this scene. In the given scene, we randomly select an input view as a pseudo test view and use the other input views to render the pseudo test view for training. Additionally, we add trainable parameters of NeuRay and a consistency loss in finetuning.

\textbf{Trainable parameters of NeuRay}. Fig.~\ref{fig:ft} shows the detailed structure of the CNN in Fig.~\ref{fig:gen_pipeline}. On every input view, an intermediate feature map $\bm{G}'\in \mathbb{R}^{H\times W \times C}$ between the constructed cost volume (or the estimated depth map) and the visibility feature map $\bm{G}$ is treated as trainable parameters of NeuRay for this input view. We call the convolution layers before the $\bm{G}'$ as an \textit{initialization network} and the convolution layers between $\bm{G}'$ and $\bm{G}$ as a \textit{visibility encoder}. Parameters $\bm{G}'$ are not trained from scratch but initialized by the initialization network using the constructed cost volume or the estimated depth map. Then, the initialization network is discarded while $\bm{G}'$ along with other network parameters are optimized in finetuning. 

\textbf{Discussion}. Alternatively, we may directly make $\bm{G}$ as trainable parameters of NeuRay and discard both the initialization network and the visibility encoder. However, we find that making $\bm{G}'$ trainable and applying the visibility encoder on the trainable $\bm{G}'$ improve the visibility prediction. 
Because the convolution layers in the visibility encoder associate feature vectors of nearby pixels on $\bm{G}'$ and these nearby pixels usually have similar visibility.

\textbf{Consistency loss}. Besides the rendering loss, we additionally use a consistency loss in finetuning. Since both the constructed radiance field and the visibility of NeuRay depict the scene geometry, we can enforce the consistency between them in finetuning.
Specifically, in finetuning, we sample points $\bm{p}_i\equiv \bm{p}(z_i)$ on a pseudo test ray and compute the hitting probability $h_i$ on the sample points from the constructed radiance field. 
Meanwhile, the pseudo test view is also an input view, on which there is a NeuRay representation to decode a distribution $t(z)$ for this pseudo test ray.
Based on $t(z)$, we compute a new hitting probability $\tilde{h}_i$ on every sample point $\bm{p}_i$ by
\begin{equation}
    \tilde{h}_i = t(z_{i+1}) - t(z_i),
    \label{eq:hit_input}
\end{equation}
where $z_i$ is the depth of the point.
Thus, we can enforce the consistency between $\tilde{h}_i$ and $h_i$ to construct a loss
\begin{equation}
    \ell_{consist}=\frac{1}{K_t}\sum_{i=1}^{K_t} CE(\tilde{h}_i,h_i),
\end{equation}
where $CE$ is the cross entropy loss.

\textbf{Discussion}. Similar to previous generalization methods~\cite{wang2021ibrnet,Yu20arxiv_pixelNeRF,chibane2021stereo,Trevithick20arxiv_GRF}, finetuning on a specific scene refines our network parameters for better feature aggregation and thus better radiance field construction on the scene. Meanwhile, further adding the trainable parameters of NeuRay and the consistency loss enables our rendering framework to refine the NeuRay representation, which brings better occlusion inference and significantly improves the rendering quality. This actually enables a memorization mechanism as illustrated in the appendix.

\subsection{Speeding up rendering with NeuRay}
\label{sec:speed}
On a test ray, most sample points $\bm{p}_i$ have nearly zero hitting probabilities $h_i$ and thus do not affect the output color. However, most computations are wasted on feature aggregation on these empty points. In the following, we show that a coarse hitting probability $\hat{h}_i$ on a test ray can be directly computed from NeuRay using little computation. 
Then, only very few fine points are sampled around points with large $\hat{h}_i$ for feature aggregation.

To compute the hitting probability $\hat{h}_i$ defined on the \textbf{test ray}, we first define the alpha value $\tilde{\alpha}$ in depth range $(z_0,z_1)$ on an \textbf{input ray} by
\begin{equation}
    \tilde{\alpha}(z_0,z_1)=\frac{t(z_1)-t(z_0)}{1-t(z_0)}.
    \label{eq:alpha_input}
\end{equation}
Then, we compute the $\hat{h}_i$ on a sample point $\bm{p}_i$ by
\begin{equation}
    \hat{\alpha}_{i}=\frac{\sum_{j}\tilde{\alpha}_{i,j}(z_{i,j},z_{i,j}+l_i) v_{i,j}}{\sum_{j} v_{i,j}},
    \label{eq:que_exist}
\end{equation}
\begin{equation}
    \hat{h}_{i}=\prod_{k=1}^{i-1} (1-\hat{\alpha}_k)\hat{\alpha}_i,
\end{equation}
where $z_{i,j}$ is the depth of the point on $j$-th input view, $l_i=z_{i+1}-z_i$ is the distance between the point $\bm{p}_i$ and its subsequent point $\bm{p}_{i+1}$ on the test ray. Computation of $\hat{h}$ is very fast because it only involves a simple combination of $t(z)$ on input views. We discuss the rationale behind the design of $\tilde{\alpha}$, $\hat{\alpha}$ and $\hat{h}$ and their connections to NeRF\cite{mildenhall2020nerf}-style density in the appendix.

\begin{table*}[ht]
    \centering
    \resizebox{\textwidth}{!}{
    \begin{tabular}{clccccccccc}
    \toprule
             &        & \multicolumn{3}{c}{Synthetic Object NeRF} & \multicolumn{3}{c}{Real Object DTU} & \multicolumn{3}{c}{Real Forward-facing LLFF} \\
     Settings & Method & PSNR$\uparrow$ & SSIM$\uparrow$ & LPIPS$\downarrow$ & PSNR$\uparrow$ & SSIM$\uparrow$ & LPIPS$\downarrow$ & PSNR$\uparrow$ & SSIM$\uparrow$ & LPIPS$\downarrow$ \\
    \midrule
    % & LLFF~\cite{mildenhall2019local}
    % & 24.77 & 0.911 & 0.114 &    -  &     - &     - & 24.41 & 0.805 & 0.211 \\
    \multirow{4}{*}{Generalization}
    & PixelNeRF~\cite{Yu20arxiv_pixelNeRF} 
    & 22.65 & 0.808 & 0.202 & 19.40 & 0.463 & 0.447 & 18.66 & 0.588 & 0.463 \\
    & MVSNeRF~\cite{chen2021mvsnerf}
    & 25.15 & 0.853 & 0.159 & 23.83 & 0.723 & 0.286 & 21.18 & 0.691 & 0.301 \\
    & IBRNet~\cite{wang2021ibrnet}
    & 26.73 & 0.908 & 0.101 & 25.76 & 0.861 & 0.173 & 25.17 & 0.813 & 0.200 \\
    % & Ours      & \textbf{28.91} & \textbf{0.920} & \textbf{0.095} & \textbf{28.30} & \textbf{0.907} & \textbf{0.130} & \textbf{25.85} & \textbf{0.832} & \textbf{0.190} \\
    & Ours      
    & \textbf{28.29} & \textbf{0.927} & \textbf{0.080} &  \textbf{26.47} & \textbf{0.875} & \textbf{0.158} & \textbf{25.35} & \textbf{0.818} & \textbf{0.198}  \\
    \midrule
    \multirow{4}{*}{Finetuning}
    & MVSNeRF~\cite{chen2021mvsnerf}
    &  27.21 & 0.888 & 0.162 & 25.41 & 0.767 & 0.275 & 23.54 & 0.733 & 0.317 \\
    & NeRF~\cite{mildenhall2020nerf}
    & 31.01 & 0.947 & 0.081 & 28.11 & 0.860 & 0.207 & 26.74 & 0.840 & 0.178 \\
    & IBRNet~\cite{wang2021ibrnet}
    & 30.05 & 0.935 & 0.066 & 29.17 & 0.908 & 0.128 & 26.87 & 0.848 & 0.175  \\
    & Ours      & \textbf{32.35} & \textbf{0.960} & \textbf{0.048} & \textbf{29.79}  & \textbf{0.928} & \textbf{0.107} &  \textbf{27.06} & \textbf{0.850} & \textbf{0.172} \\
    \bottomrule
    \end{tabular}
    }
    \caption{Quantitative comparison with baseline methods.}
    \vspace{-10pt}
    \label{tab:results}
\end{table*}
\section{Experiment}

\subsection{Experimental Protocols}
\subsubsection{Datasets}
We use two kinds of evaluation datasets, the object dataset and the forward-facing dataset. The results are evaluated by PSNR, SSIM~\cite{hore2010image} and LPIPS~\cite{zhang2018unreasonable} as the metrics.

\newcommand{\resultsize}{0.113}
\begin{figure*}
    \centering
    \setlength\tabcolsep{1.5pt}
    \begin{tabular}{cc|ccc|ccc}
               &    & \multicolumn{3}{c|}{Generalization}   &\multicolumn{3}{c}{Finetuning} \\
         Image & GT & MVSNeRF & IBRNet &  NeuRay & MVSNeRF & IBRNet & NeuRay\\
         \includegraphics[width=\resultsize\textwidth]{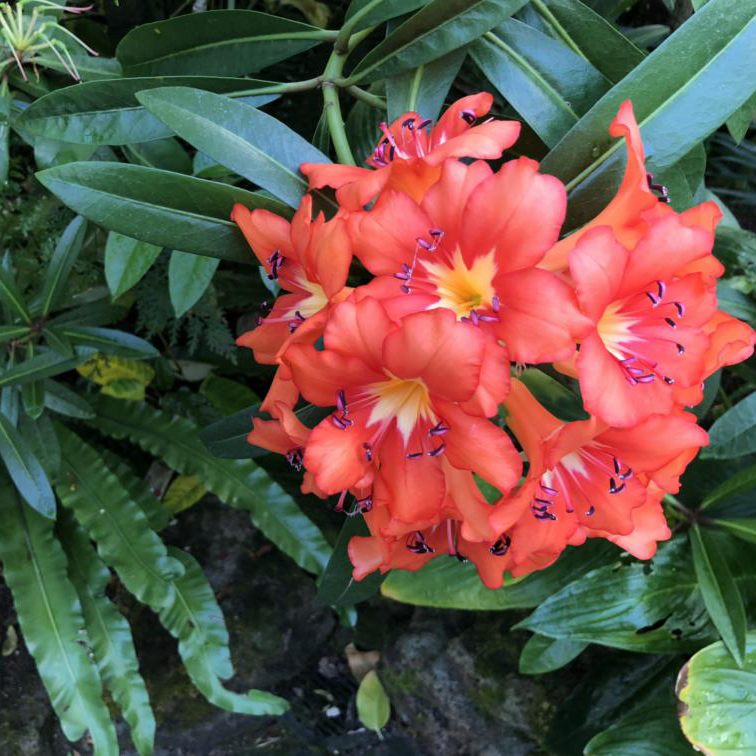} &
         \includegraphics[width=\resultsize\textwidth]{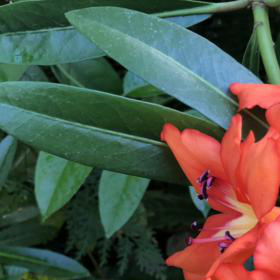} &
         \includegraphics[width=\resultsize\textwidth]{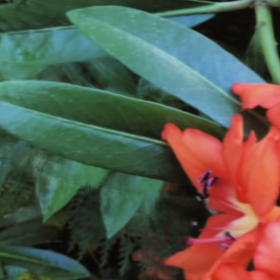} &
         \includegraphics[width=\resultsize\textwidth]{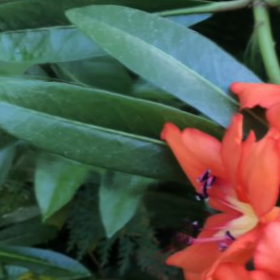} &
         \includegraphics[width=\resultsize\textwidth]{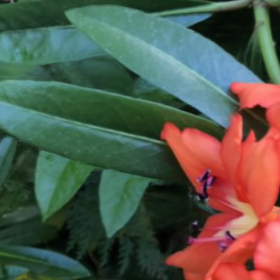} &
         \includegraphics[width=\resultsize\textwidth]{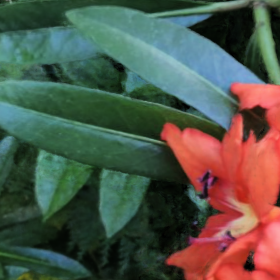} &
         \includegraphics[width=\resultsize\textwidth]{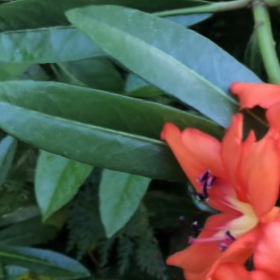} &
         \includegraphics[width=\resultsize\textwidth]{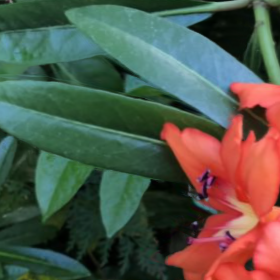} \\
         
         \includegraphics[width=\resultsize\textwidth]{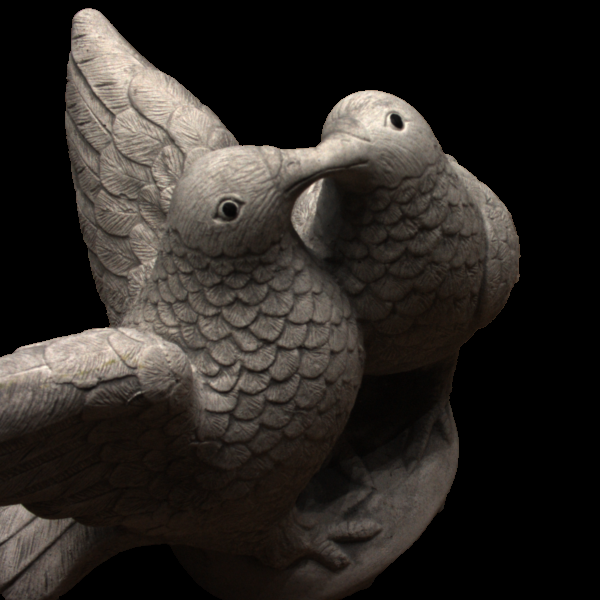} &
         \includegraphics[width=\resultsize\textwidth]{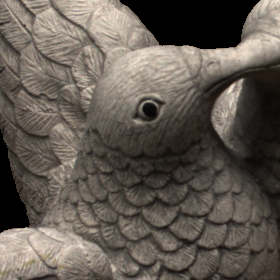} &
         \includegraphics[width=\resultsize\textwidth]{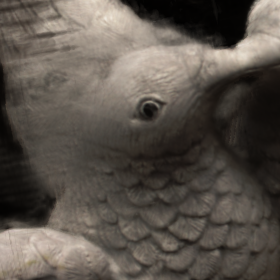} &
         \includegraphics[width=\resultsize\textwidth]{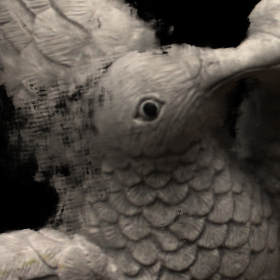} &
         \includegraphics[width=\resultsize\textwidth]{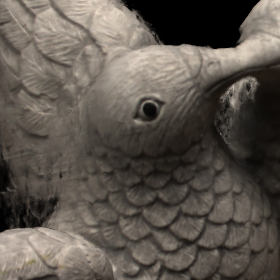} &
         \includegraphics[width=\resultsize\textwidth]{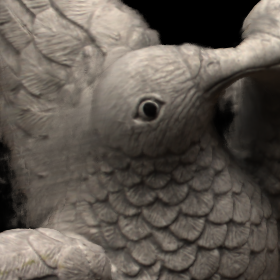} &
         \includegraphics[width=\resultsize\textwidth]{image_new/results/birds_ibrnet_ft_crop.png} &
         \includegraphics[width=\resultsize\textwidth]{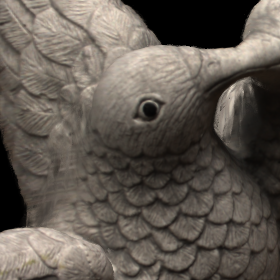} \\

         \includegraphics[width=\resultsize\textwidth]{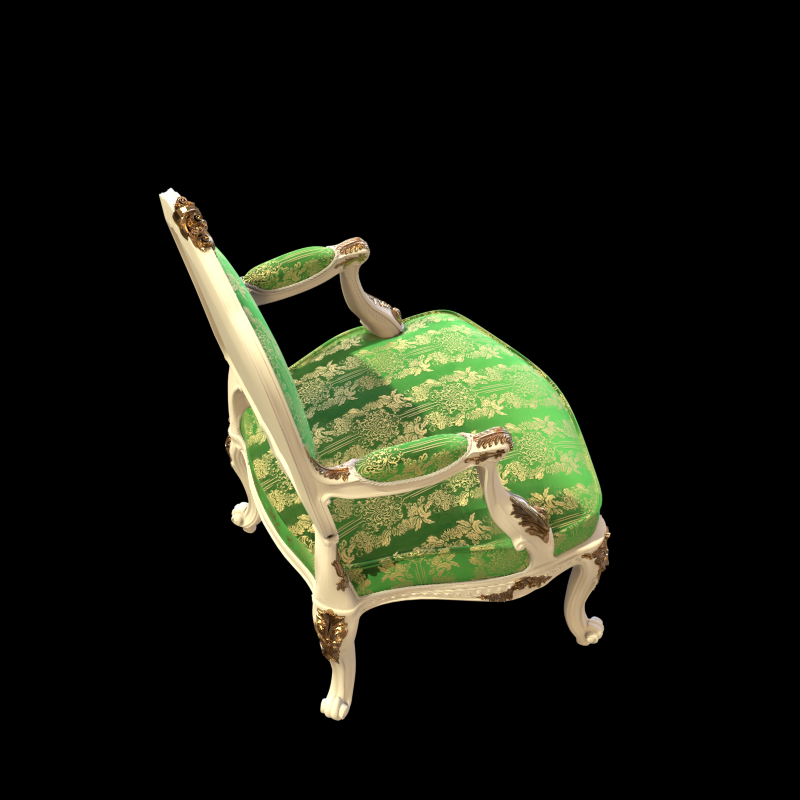} &
         \includegraphics[width=\resultsize\textwidth]{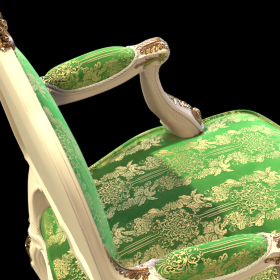} &
         \includegraphics[width=\resultsize\textwidth]{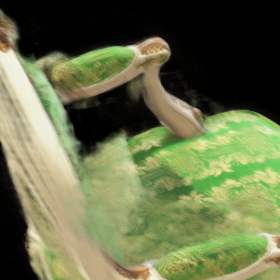} &
         \includegraphics[width=\resultsize\textwidth]{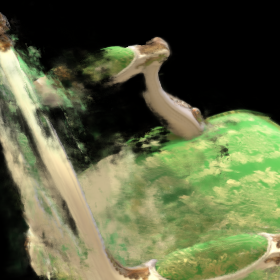} &
         \includegraphics[width=\resultsize\textwidth]{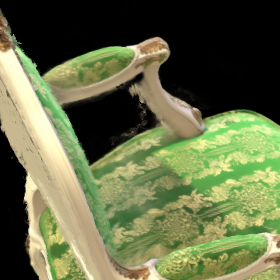} &
         \includegraphics[width=\resultsize\textwidth]{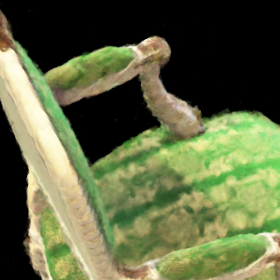}&
         \includegraphics[width=\resultsize\textwidth]{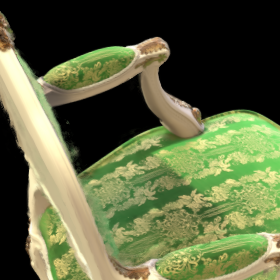} &
         \includegraphics[width=\resultsize\textwidth]{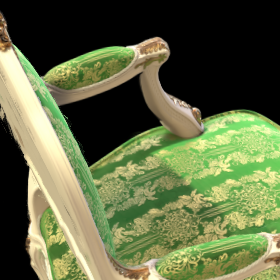} \\
    \end{tabular}
    \caption{Qualitative results of different methods. Please refer to the appendix for more results.}
    \vspace{-10pt}
    \label{fig:results}
\end{figure*}
\textbf{Object dataset}. The object dataset includes the NeRF synthetic dataset \cite{niemeyer2020differentiable} and the DTU dataset~\cite{jensen2014large}. The NeRF synthetic dataset has 8 objects, each of which contains 100 images as input views and the other 200 images as test views. For the DTU dataset, we select 4 objects (birds, tools, bricks and snowman) as test objects. On each test object, we leave out 1/8 images as test views and the rest images as input views.

\textbf{Forward-facing dataset}. The forward-facing dataset is the LLFF dataset~\cite{mildenhall2019local} with 8 scenes. Each scene contains 20 to 62 images. We follow the same train-test set split for each scene as previous methods~\cite{mildenhall2020nerf,wang2021ibrnet}, which uses 1/8 images as test views. As done in \cite{Wizadwongsa2021NeX}, all images are undistorted by COLMAP~\cite{schoenberger2016sfm}. The evaluation resolution are 1008$\times$756 for the LLFF dataset, 800$\times$800 for the NeRF synthetic dataset and 800$\times$600 for the DTU dataset. The test images in two object datasets all use black backgrounds. 

\textbf{Training dataset}. In order to train the generalization model, we use three kinds of datasets: (1) the synthetic Google Scanned Object dataset~\cite{google_scanned_objects}, which contains 1023 objects with 250 rendered images on each object; (2) three forward-facing training datasets~\cite{zhou2018stereo,mildenhall2019local,flynn2019deepview} and (3) the rest training objects from the DTU dataset.

\begin{figure}
    \centering
    \setlength\tabcolsep{1.5pt}
    \begin{tabular}{cc|cc}
    \multicolumn{2}{c|}{NeRF\cite{mildenhall2020nerf}} & \multicolumn{2}{c}{NeuRay-Ft} \\
    \includegraphics[width=0.22\linewidth]{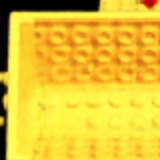} &
    \includegraphics[width=0.22\linewidth]{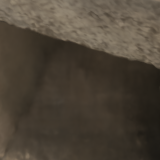} &
    \includegraphics[width=0.22\linewidth]{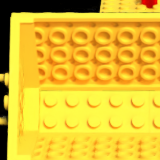} &
    \includegraphics[width=0.22\linewidth]{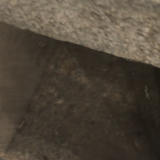} \\
    \end{tabular}
    \caption{Our method renders details more clearly than NeRF~\cite{mildenhall2020nerf}.}
    \vspace{-10pt}
    \label{fig:nerf_com}
\end{figure}
\begin{figure}
    \centering
    \begin{tabular}{ccc}
         GT & Single & Mixture \\
         \includegraphics[width=0.25\linewidth]{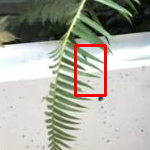} &
         \includegraphics[width=0.25\linewidth]{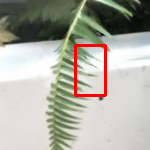} &
         \includegraphics[width=0.25\linewidth]{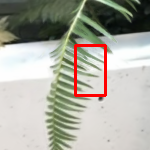} \\
    \end{tabular}
    \caption{Comparison between the mixture logistics distribution $N_l=2$ and the single logistics distribution $N_l=1$ as the occlusion probability. Mixture logistics distribution improves the results on the edges with abrupt depth changes ({\color{red} red rectangle}).}
    \vspace{-10pt}
    \label{fig:mixture}
\end{figure}

\vspace{-10pt}
\subsubsection{Implementation details} 
To render a test view, we do not use all input views but $N_w=8$ neighboring input views, called \textbf{working views}. $D=64$ planes are used in the cost volume while we use COLMAP~\cite{schoenberger2016mvs} for patch-match stereo. $N_l=2$ logistics distributions are mixed in $t(z)$. We use the coarse-to-fine sampling strategy as done in \cite{mildenhall2020nerf, wang2021ibrnet} with 64 sample points in both stages. Coarse and fine models share the same image encoder, visibility encoder and initialization network, but they use different decoder $\mathcal{F}$ and aggregation networks. The aggregation networks follows similar design as \cite{wang2021ibrnet} but with additional visibility as inputs, which blends input colors and applies a transformer along test rays. All experiments are conducted on a 2080 Ti GPU. Details and architectures can be found in the appendix.

\newcommand{\convsize}{0.14}
\begin{figure*}
    \centering
    \setlength\tabcolsep{1.5pt}
    \begin{tabular}{cccccc}
    \multirow{2}{*}{\includegraphics[width=0.25\textwidth]{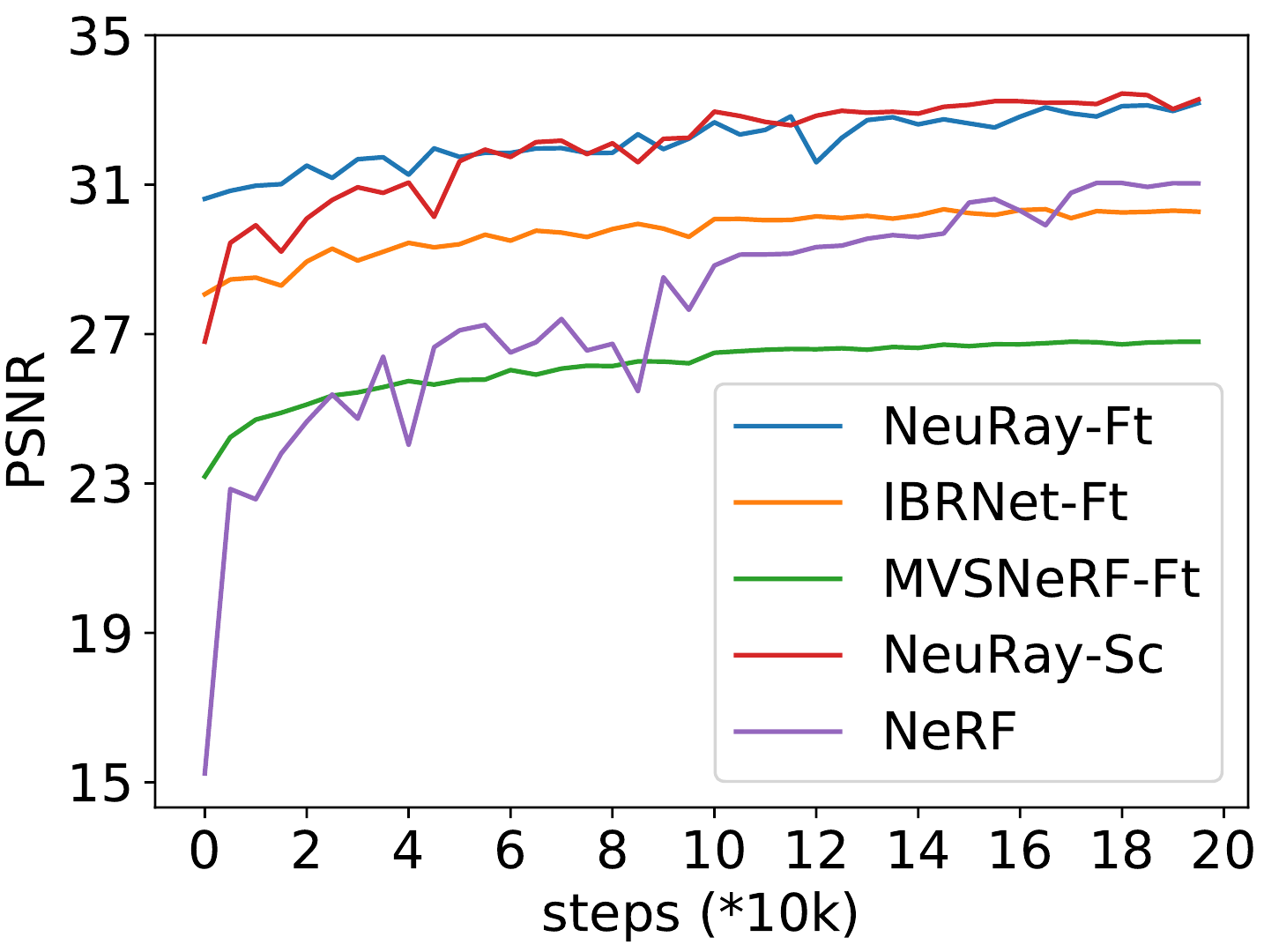}} 
    & GT & NeRF~\cite{mildenhall2020nerf} & MVSNeRF~\cite{chen2021mvsnerf} & IBRNet~\cite{wang2021ibrnet} & NeuRay-Ft \\ & 
    \includegraphics[width=\convsize\textwidth]{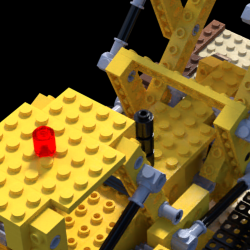}& 
    \includegraphics[width=\convsize\textwidth]{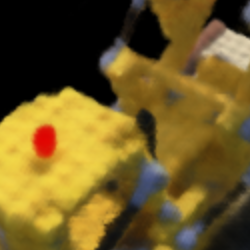}& 
    \includegraphics[width=\convsize\textwidth]{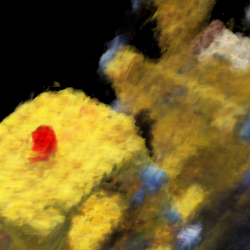}& 
    \includegraphics[width=\convsize\textwidth]{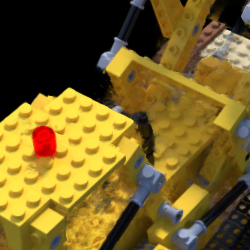}& 
    \includegraphics[width=\convsize\textwidth]{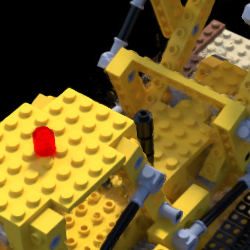}
    \\
    \end{tabular}
    \caption{ (Left) Curves of PSNR of different models with different training steps. PSNR is computed on a validation set of the ``Lego". (Right) Qualitative results of models with 10k training steps. Due to the limitation of GPU memory, the batch size for IBRNet~\cite{wang2021ibrnet} and our method is 512 while the batch size for MVSNeRF~\cite{chen2021mvsnerf} and NeRF~\cite{mildenhall2020nerf} is 1024.}
    \vspace{-10pt}
    \label{fig:convergence}
\end{figure*}
\subsection{Comparison with baselines}
\textbf{Experiment settings}. We compare with IBRNet~\cite{wang2021ibrnet}, PixelNeRF~\cite{Yu20arxiv_pixelNeRF}, MVSNeRF~\cite{chen2021mvsnerf} and NeRF~\cite{mildenhall2020nerf} in the generalization setting and the finetuning setting. In the generalization setting, all generalization methods including our method are pretrained on the same training scenes and tested on unseen test scenes. In the finetuning setting, all generalization methods including ours are further finetuned on the input views of each test scene while NeRF~\cite{mildenhall2020nerf} is trained-from-scratch.
% MVSNeRF~\cite{chen2021mvsnerf} needs to build a cost volume in the view frustum of one input view. In the generalization setting, we build the cost volume on the nearest input view to each query view. However, finetuning MVSNeRF has to fix the cost volume on one view, in which the input view locating at the center is used. 

The quantitative results are shown in Table.~\ref{tab:results} and the qualitative results are shown in Fig.~\ref{fig:results}. 
Table~\ref{tab:results} shows that our method generalizes well to unseen scenes and outperforms all other generalization models. 
After scene-specific finetuning, our method clearly outperforms all baselines on two object datasets but achieves similar performance as IBRNet~\cite{wang2021ibrnet} on the LLFF dataset~\cite{mildenhall2019local}. 
The reason is that the LLFF dataset contains very dense forward-facing input views so that every 3D point is visible to a large number of input views. 
In this case, even one or two views are occluded, there are still enough visible views to provide feature consistency for IBRNet~\cite{wang2021ibrnet} to render correctly. 
In comparison, in the Synthetic NeRF dataset, images are sparsely captured around the object in 360$^\circ$, which brings more severe feature inconsistency to reduce the performance of IBRNet~\cite{wang2021ibrnet}. 
In contrast, our rendering is occlusion-aware so that our model performs much better on the NeRF synthetic dataset. Furthermore, we show that our method outperforms IBRNet by a large margin with sparse working views on the LLFF dataset in the appendix.

By comparing MVSNeRF~\cite{chen2021mvsnerf} with our method in Fig.~\ref{fig:results}, we notice that finetuning MVSNeRF~\cite{chen2021mvsnerf} leads to noisy artifacts when the test view is far from the input view on which the cost volume is built.
By comparing NeRF~\cite{mildenhall2019local} with our method in Fig.~\ref{fig:nerf_com}, we find that NeRF needs more optimization steps to recover subtle details like surfaces of the bricks and textures on the Lego while it is relatively more easy for our method to render these details by blending colors of input views, which is the reason that our method can achieve better rendering quality than NeRF.
  
% Meanwhile,  we find that undistortion by Colmap~\cite{schoenberger2016mvs} is not perfect and noticeable distortions still remain, which makes the scene-specific optimization struggle to improve. Fig~\ref{fig:results} shows that NeuRay improves the rendering quality in the occlusion regions. 

\subsection{Ablation studies}
% We conduct ablation studies on the ``Lego" from the NeRF synthetic dataset and the ``Fern" from the LLFF dataset. Results in PSNR are reported in Table~\ref{tab:ablation}.

\textbf{How effective is the visibility from init-NeuRay?}
To validate this, in Table~\ref{tab:ablation}, we test the IBRNet~\cite{wang2021ibrnet} (ID 1),  the model with only image feature aggregation (ID 2), the model aggregating images features with estimated depth from COLMAP~\cite{schoenberger2016mvs} (ID 3) and the model aggregating image features with visibility of NeuRay initialized by cost volumes (ID 4). Since our image encoder is shallower than IBRNet, the performance with only image feature aggregation is worse than IBRNet. Simply adding estimated depth only brings slight improvement while using visibility of initialized NeuRay significantly improves the quality.

\textbf{Can NeuRay be trained from scratch?}
To show NeuRay can be constructed on a scene from scratch without initialization from cost volumes, we follow exactly the same process as finetuning to train our method but the raw visibility feature maps $\bm{G}'$ and parameters of all networks are randomly initialized. Results in Table~\ref{tab:ablation} (ID 5 and 6) show that training our method from scratch is also able to achieve similar results as finetuning the initialized pretrain model.

\begin{table}[]
    \centering
    \resizebox{\linewidth}{!}{
    \begin{tabular}{ccccc}
        \toprule
        ID & Description & Setting & Lego & Fern \\
        \midrule
        1 & IBRNet                                & Gen &  25.64 & 24.16  \\
        2 & Only aggregation               & Gen &  25.61 & 22.25 \\
        3 & Aggregation with depth features & Gen &  26.45 & 22.43 \\
        % 4 & $M,B,P$ with init-NeuRay-D & Gen &  29.79 & 24.69 \\
        4 & Aggregation with init-NeuRay    & Gen & 28.41 & 24.02 \\
        \midrule
        5 & mixture logistics $N_l=2$& Ft & 32.97 & 25.93 \\
        6 & mixture logistics $N_l=2$& Sc & 33.07 & 25.89 \\
        7 & single logistics $N_l=1$ & Sc & 33.05 & 25.58 \\
        8 & Only aggregation   & Sc & 29.61 & 24.40 \\
        9 & NeuRay without $\ell_{consist}$ & Sc & 31.46 & 25.24 \\
        \bottomrule
    \end{tabular}
    }
    \caption{Ablation studies. PSNRs on the ``Lego" from the NeRF synthetic dataset and the ``Fern" from the LLFF dataset are reported. ``Gen" means the generalization setting, ``Ft" means finetuning on the scene and ``Sc" means training from scratch.}
    \vspace{-15pt}
    \label{tab:ablation}
\end{table}
\begin{table}[]
    \centering
    \begin{tabular}{cccccc}
    \toprule
         Method & $\hat{h}$ & $K_{t,1}$ & $K_{t,2}$  & PSNR  & Time(s) \\
         \midrule
         IBRNet & \xmark    &   64      &    64     & 34.00 &  31.46  \\
         NeRF   & \xmark    &   128     &   128     & 33.65 &  31.51  \\
         NeuRay & \xmark    &   64      &   64      & 35.33 &  30.03  \\
         NeuRay & \cmark    &   64      &    8      & 34.57 &   3.95  \\ % 8_8_prob
        %  NeuRay & \cmark    &   32      &    8      &    8  & 34.10 &   3.46  \\ % 8_8_prob_v5_v2
         NeuRay & \cmark    &   32      &    4      & 33.73 &   2.57  \\ % 8_8_prob_v5
        %  NeuRay & \cmark    &   32      &    4      &    4  &       &         \\
         \bottomrule
    \end{tabular}
    \caption{Rendering time and PSNR of the ``birds" from the DTU dataset. $K_{t,1}$ and $K_{t,2}$ are numbers of points used in coarse and fine sampling respectively, $\hat{h}$ means using the probability $\hat{h}$ to conduct coarse sampling, ``Time" means the time cost on rendering one 800$\times$600 images on a 2080Ti GPU.}
    \label{tab:speed}
\end{table}
\begin{table}[]
    \centering
    \resizebox{\linewidth}{!}{
    \begin{tabular}{ccccc}
        \toprule
         Method & Train Step & Train Time & PSNR \\
         \midrule
        %  NeRF~\cite{mildenhall2020nerf} & 10k  & $\sim$30min &       \\
         NeRF~\cite{mildenhall2020nerf}    & 200k & $\sim$9.5h  & 30.27 \\
         MVSNeRF~\cite{chen2021mvsnerf}-Ft & 10k  & $\sim$28min & 23.77 \\
         IBRNet~\cite{wang2021ibrnet}-Ft   & 5k   & $\sim$41min & 28.38 \\
         NeuRay-Ft                         & 5k   & $\sim$32min & 30.63 \\
         \bottomrule
    \end{tabular}
    }
    \caption{PSNR and training steps/time on NeRF synthetic dataset.}
    \vspace{-10pt}
    \label{tab:low_res}
\end{table}
\textbf{Single or mixture logistics distributions?}
As discussed in Sec.~\ref{sec:neuray}, the choice of $t(z)$ can be a single logistics distribution or a mixture of logistics distributions. We compare these two choices in Table~\ref{tab:ablation} (ID 6 and 7) and show qualitative comparison in Fig.~\ref{fig:mixture}, which demonstrates that using a mixture of logistics distributions improves the rendering quality in regions with abrupt depth changes.

\textbf{How effective is NeuRay in per-scene optimization?}
In Table~\ref{tab:ablation}, we compare three models, the full model (ID 6), the model with only image feature aggregation (ID 8) and the model with NeuRay but without $\ell_{consist}$ (ID 9). The results show that adding a NeuRay as a backend in per-scene optimization already brings improvements and further adding $\ell_{consist}$ enables memorization of geometry thus greatly improves the rendering quality.

\subsection{Analysis}
\textbf{Speed up rendering with NeuRay}. As discussed in Sec.~\ref{sec:speed}, we can efficiently estimate a coarse $\hat{h}$ from the NeuRay to perform the coarse sampling and only sample few fine points based on the coarse sampling. To validate this, we conduct an experiment on the ``birds" from the DTU dataset~\cite{jensen2014large}. As shown in Table~\ref{tab:speed}, only 4 or 8 subsequent sample points are enough for our method to achieve high-quality renderings, which speeds up the rendering 10 times from 30s to  $\sim$3s. Further speeding up by baking out a mesh or occupancy voxels from alpha values $\hat{e}$ is possible, which we leave for future works.
% To further speed up the computation of probability $t(z)$, we compute all distribution parameters $\mu_i$, $\sigma_i$ and $w_i$ on reference views beforehand once and directly interpolate on these distribution parameters to avoid querying network $\phi$. 

\textbf{Convergence speed}.
To show how the rendering quality of different models improves in the scene-specific optimization process, we train different models on the ``Lego" and plot the curves of PSNR on a small validation set in Fig.~\ref{fig:convergence} (left). The curves show that finetuning our method produces consistently better rendering results than training all baseline methods with the same training steps. In Fig.~\ref{fig:convergence} (right), we show the qualitative results on 10k training steps. With only 10k training steps, NeRF~\cite{mildenhall2020nerf} and MVSNeRF~\cite{chen2021mvsnerf} are still far from convergence thus produce blurred images, IBRNet~\cite{wang2021ibrnet} produces artifacts on regions with occlusions while our method already produces high-quality renderings. 

\textbf{Only finetune few steps with NeuRay}. Table~\ref{tab:low_res} reports PSNR and time of different models with only few finetuning steps on the NeRF synthetic dataset. Note that finetuning both IBRNet~\cite{wang2021ibrnet} and our method requires image feature extraction which costs more time on one training step than neural fields MVSNeRF~\cite{chen2021mvsnerf} and NeRF~\cite{mildenhall2020nerf}. Since our image encoder is shallower than IBRNet, finetuning our method is slightly faster. The results show that our method is able to be finetuned limited time (32min) to achieve similar quality as NeRF~\cite{mildenhall2020nerf} with long training time (9.5h), which is significantly better than the other generalization methods with similar finetuning time.
% which demonstrates that finetuning NeuRay with 10k steps (40min) could produce comparable results as NeRF~\cite{mildenhall2020nerf} trained with 200k steps (9.5h).
% The plots show that IBRNet only achieves improvement on the first 10k steps and its performance even degrades slightly on the validation set with more training steps due to overfitting. 
% In contrast, NeuRay consistently improves the performance with more finetuning steps. The reason is that IBRNet lacks a memorization mechanism as explained in the memory interpretation in Sec.~\ref{sec:spe_nr}.

\vspace{-5pt}
\section{Conclusion}
In this paper, we proposed a novel neural representation NeuRay for the novel view synthesis task. NeuRay represents a scene by occlusion probabilities defined on input rays and is able to efficiently estimate the visibility from arbitrary 3D points to input views. With the help of NeuRay, we are able to consider the visibility when constructing radiance fields by multi-view feature aggregation. Experiments on the DTU dataset, the NeRF synthetic dataset and the LLFF dataset demonstrate that our method can render high-quality images without any training on the scene or with only few finetuning steps on the scene.

%%%%%%%%% REFERENCES
{\small
\bibliographystyle{ieee_fullname}
\bibliography{egbib}
}

\section*{Appendices}
\setcounter{equation}{0}
\setcounter{subsection}{0}
\renewcommand{\theequation}{A.\arabic{equation}}
\renewcommand{\thesubsection}{A.\arabic{subsection}}

\subsection{Comparison between NeRF~\cite{mildenhall2020nerf} and NeuRay}

\label{sec:compare}

In this section, we will compare NeRF~\cite{mildenhall2020nerf} with NeuRay on parameterizing different probabilities on a ray. Let us first review NeRF's definitions of different probabilities. 

\begin{figure}
    \centering
    \includegraphics[width=\linewidth]{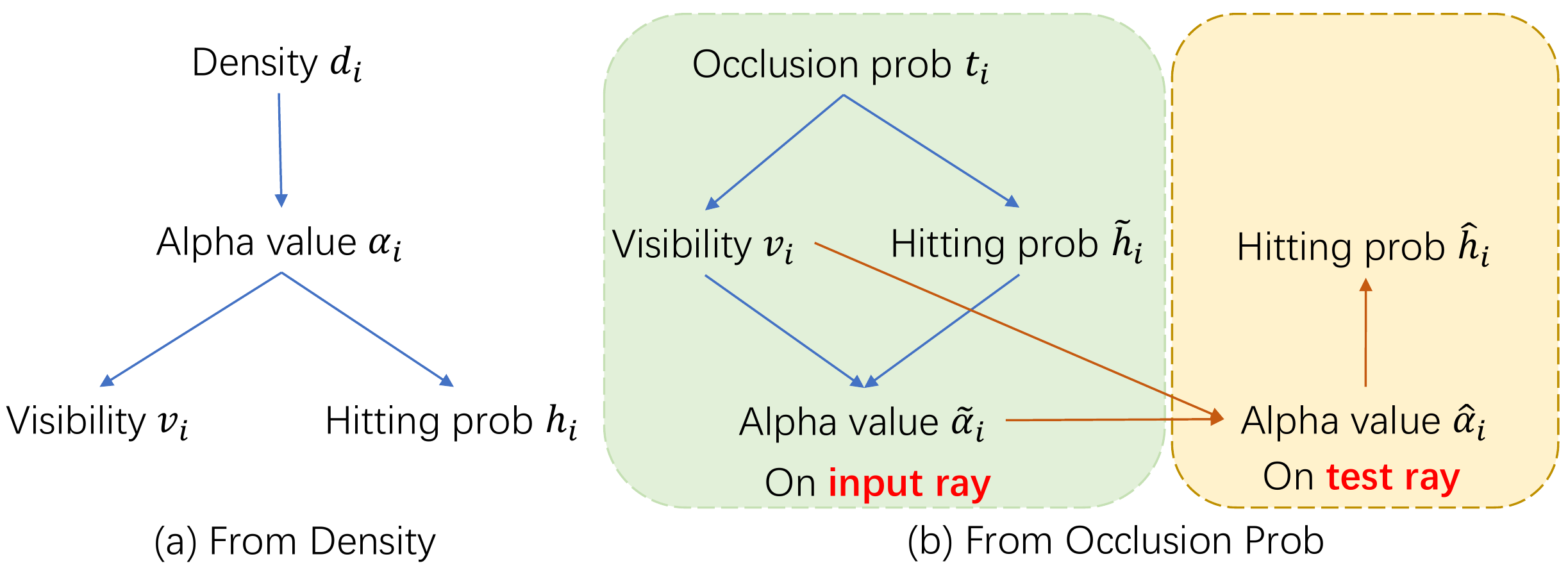}
    \caption{
    Comparison between the density in NeRF and the occlusion probability in NeuRay. (a) In NeRF, the network decodes the density to compute the alpha values, the hitting probability and the visibility. (b) In NeuRay, the network decodes the parameters of occlusion probability, which enables efficient computation of the visibility and the hitting probability. Then, we derive the corresponding alpha values from the hitting probability and visibility.}
    \label{fig:discuss}
\end{figure}

\textbf{NeRF's parameterization}. Given a ray $\bm{p}(z) =\bm{o}+z\bm{r}$ with $\{ \bm{p}_i\equiv \bm{p}(z_i)| i=1,...,N; z_{i}<z_{i+1}; z_i\in \mathbb{R}^{+} \}$ sample points on the ray, NeRF computes the densities of these points by $d_{i}=\mathcal{F}(\bm{p}_i)$, where $\mathcal{F}$ is an MLP. Then, the alpha values on these points are computed by $\alpha_{i}=1-\exp(-{\rm ReLU}(d_i) l_i)$, where $l_i=z_{i+1}-z_i$. In this case, the visibility from the ray origin $\bm{o}$ to the point $\bm{p}_i$ is 
\begin{equation}
    v_i=\prod_{j=1}^{i-1} (1-\alpha_j).
\end{equation}
The hitting probability, which means the ray is not occluded by any depth up to $z_i$ and hits a surface in the range $(z_i,z_{i+1})$, is
\begin{equation}
    h_i=v_i \alpha_i
    \label{eq:hit}
\end{equation}
This process is summarized by Fig.~\ref{fig:discuss} (a), where all probabilities are based on the density $d_i$.

\textbf{NeuRay's parameterization}. In comparison, all probabilities in NeuRay are based on the occlusion probability $t(z)$. Given an \textbf{input ray}, we also sample points $\{ \bm{p}_i\equiv \bm{p}(z_i)| i=1,...,N; z_{i}<z_{i+1}; z_i\in \mathbb{R}^{+} \}$. Based on $t(z)$, we compute the hitting probability  $\tilde{h}_i$ (Eq.~(\ref{eq:hit_input}) in the main paper) and the visibility $v_i$
\begin{equation}
    \tilde{h}_i = t(z_{i+1})-t(z_i),
\end{equation}
\begin{equation}
    v_i \equiv v(z_{i})= 1-t(z_i).
\end{equation}
We use $\tilde{h}$ here to indicate it was defined on an \textit{input ray}. 
Then, noticing the relationship between the visibility, the hitting probability and the alpha value in Eq.\eqref{eq:hit}, we can compute the alpha value $\tilde{\alpha}_i$ on $\bm{p}_i$ from $\tilde{h}_i$ and $v_i$ by $\tilde{\alpha}_i= \tilde{h}_i/v_i = (t(z_{i+1})-t(z_i))/(1-t(z_i))$, which is the Eq.~(\ref{eq:alpha_input}) in the main paper. $\tilde{\alpha}$ also means it is defined on an \textit{input ray}.
This process is summarized by Fig.~\ref{fig:discuss} (b), which is inverse to the density definition in NeRF.

\textbf{Motivation of designing the hitting probability $\hat{h}$ as the form in Sec. 3.6}. 
Our motivation is explained as follows. 
When we need to compute the hitting probability on the sample points $\bm{p}_i$ of a \textbf{test ray}, we need to derive the alpha values $\hat{\alpha}_i$ on these points and compute the hitting probability by $\hat{h}_i = \prod_{j=1}^{i-1} (1-\hat{\alpha}_j) \hat{\alpha_i}$. Note we use $\hat{\alpha}$ to indicate that it is defined on the \textit{test ray} and to distinguish it from the alpha value $\tilde{\alpha}$ defined on \textit{input rays} and the alpha value $\alpha$ computed from \textit{the constructed radiance field}. To compute $\hat{\alpha}_i$, we choose to use the weighted sum of alpha values $\tilde{\alpha}$ from all input rays with their visibility as weights
\begin{equation}
    \hat{\alpha}_{i}=\frac{\sum_{j}\tilde{\alpha}_{i,j} v_{i,j}}{\sum_{j} v_{i,j}}.
\end{equation}
This is the Eq.~(\ref{eq:que_exist}) in the main paper. By considering the visibility, only visible input rays can affect the alpha values $\hat{\alpha}$ of the point. Finally, the alpha value $\hat{\alpha}_i$ is used in the computation of $\hat{h}_i=\prod_{j=1}^{i-1} (1-\hat{\alpha}_j) \hat{\alpha_i}$.

\subsection{Discussion about visibility from density}

In the main paper, we have discussed that parameterizing the visibility with volume density is computationally impractical, because it requires $K_r$ times forward pass of the network $\mathcal{F}$ to compute visibility from one input view to one 3D point. For example, given only 1 test ray and 8 neighboring input views, we sample 128 points on this test ray. Then, to compute the visibility from every input view to every sample point, we need to consider 8$\times$128 input rays. If we further sample 32 points on every input ray to accumulate volume density for visibility computation. It will require 8$\times$128$\times$32 times evaluations of the network $\mathcal{F}$. Such computation complexity is not affordable for training on a single 2080 Ti GPU to achieve reasonable results.

\subsection{Memorization interpretation in finetuning}
\begin{figure}
    \centering
    \includegraphics[width=\linewidth]{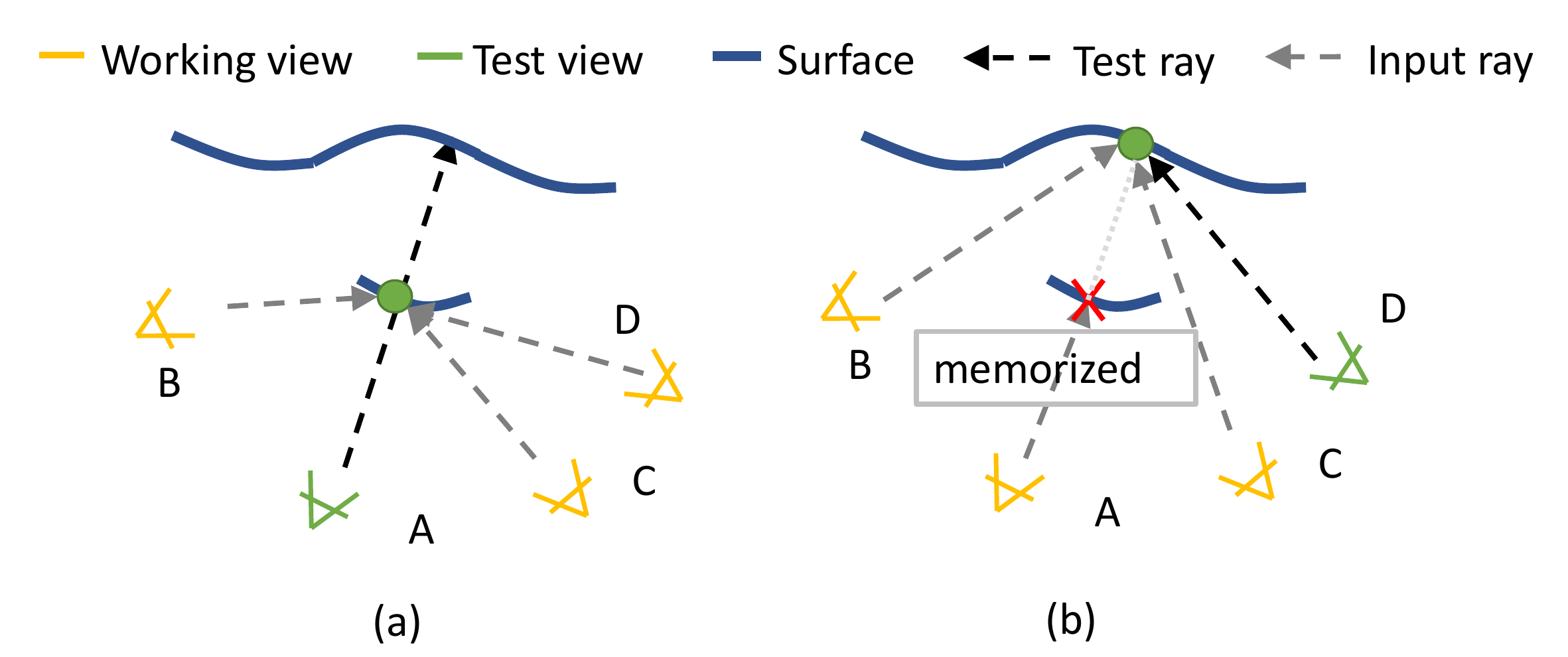}
    \caption{(a) During finetuning, when rendering the pseudo test view A using view B, C, and D, the predicted hitting probabilities and the corresponding visibility will be memorized on view A. (b) Afterwards, the memorized visibility on view A will be used for occlusion inference in rendering the pseudo test view D.}
    \label{fig:memory}
\end{figure}

$\ell_{consist}$ enables the NeuRay representation to memorize the predicted surfaces from the constructed radiance field in finetuning. An example is shown in Fig.~\ref{fig:memory}.
\vspace{-5pt}
\begin{enumerate}[leftmargin=11pt]
    \setlength{\itemsep}{1pt}
    \setlength{\parskip}{0pt}
    \setlength{\parsep}{0pt}
    \setlength{\itemindent}{0pt}
    \item In one training step as shown in Fig.~\ref{fig:memory} (a), NeuRay chooses view A as the pseudo test view and aggregates features from neighboring input views B, C and D to predict $h_{i}$ of the test ray on the test view A.
    \item  Applying the consistency loss $\ell_{consist}$ forces the visibility feature $\bm{g}$ of the test ray of A to produce $\tilde{h}_i$ consistent with $h_i$. This can be interpreted that the $h_i$ is memorized in $\bm{g}$, which also means the memorization of visibility, because both are computed from $t(z)$.
    \item  In subsequent another training step which uses view A, B and C to render the pseudo test view D in Fig.~\ref{fig:memory} (b), NeuRay already knows that the green point is invisible to A by checking the memorized visibility on A. Hence, NeuRay is able to predict an accurate $h_i$ for view D.
    \item  The predicted $h_i$ on the test ray of view D is also memorized on D and helps the prediction of hitting probabilities of other views in the subsequent training steps.
\end{enumerate}

By memorizing the scene geometry in $h_i$, NeuRay is able to refine its scene representation to provide better occlusion inference during finetuning. In contrast, existing methods~\cite{Trevithick20arxiv_GRF,wang2021ibrnet,chibane2021stereo,Yu20arxiv_pixelNeRF} can only adjust their network parameters to improve their feature aggregation during finetuning, which are still oblivious to occlusions in the scene.

\newcommand{\wnsize}{0.165}
\begin{figure*}
    \centering
    \setlength\tabcolsep{1.5pt}
    \begin{tabular}{ccc|ccc}
        IBRNet~\cite{wang2021ibrnet} & NeuRay & GT & IBRNet~\cite{wang2021ibrnet} & NeuRay & GT \\
        \includegraphics[width=\wnsize\textwidth]{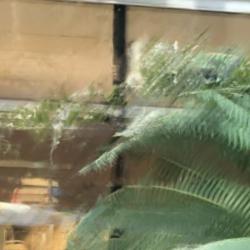}&        \includegraphics[width=\wnsize\textwidth]{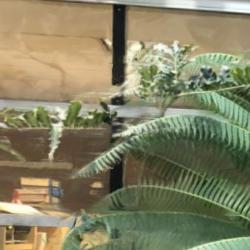}&
        \includegraphics[width=\wnsize\textwidth]{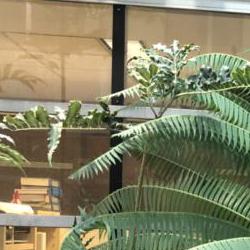}&
        \includegraphics[width=\wnsize\textwidth]{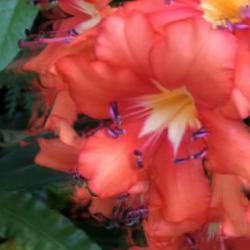}&        \includegraphics[width=\wnsize\textwidth]{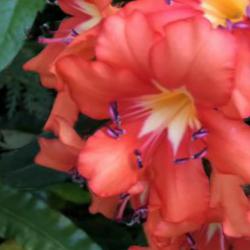}&
        \includegraphics[width=\wnsize\textwidth]{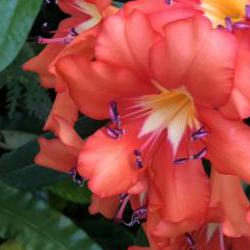}\\
        \includegraphics[width=\wnsize\textwidth]{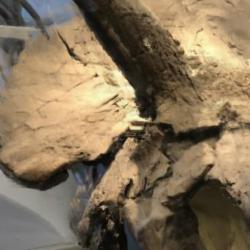}&        \includegraphics[width=\wnsize\textwidth]{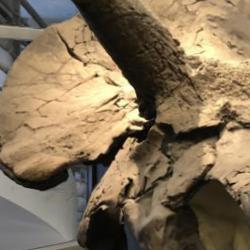}&
        \includegraphics[width=\wnsize\textwidth]{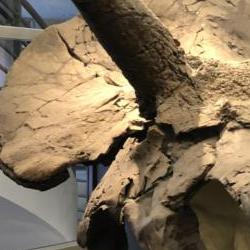}&
        \includegraphics[width=\wnsize\textwidth]{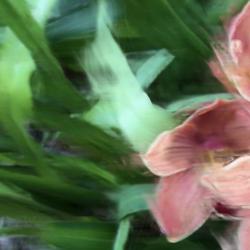}&        \includegraphics[width=\wnsize\textwidth]{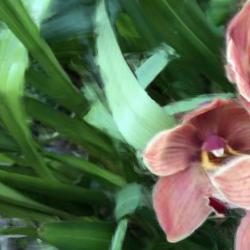}&
        \includegraphics[width=\wnsize\textwidth]{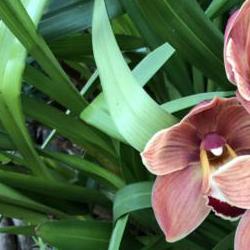}\\
    \end{tabular}
    \caption{Qualitative results using only 2 neighboring working views to render the test view.}
    \label{fig:work_num}
\end{figure*}

\subsection{Training details and architecture}
\begin{figure*}
    \centering
    \includegraphics[width=0.8\textwidth]{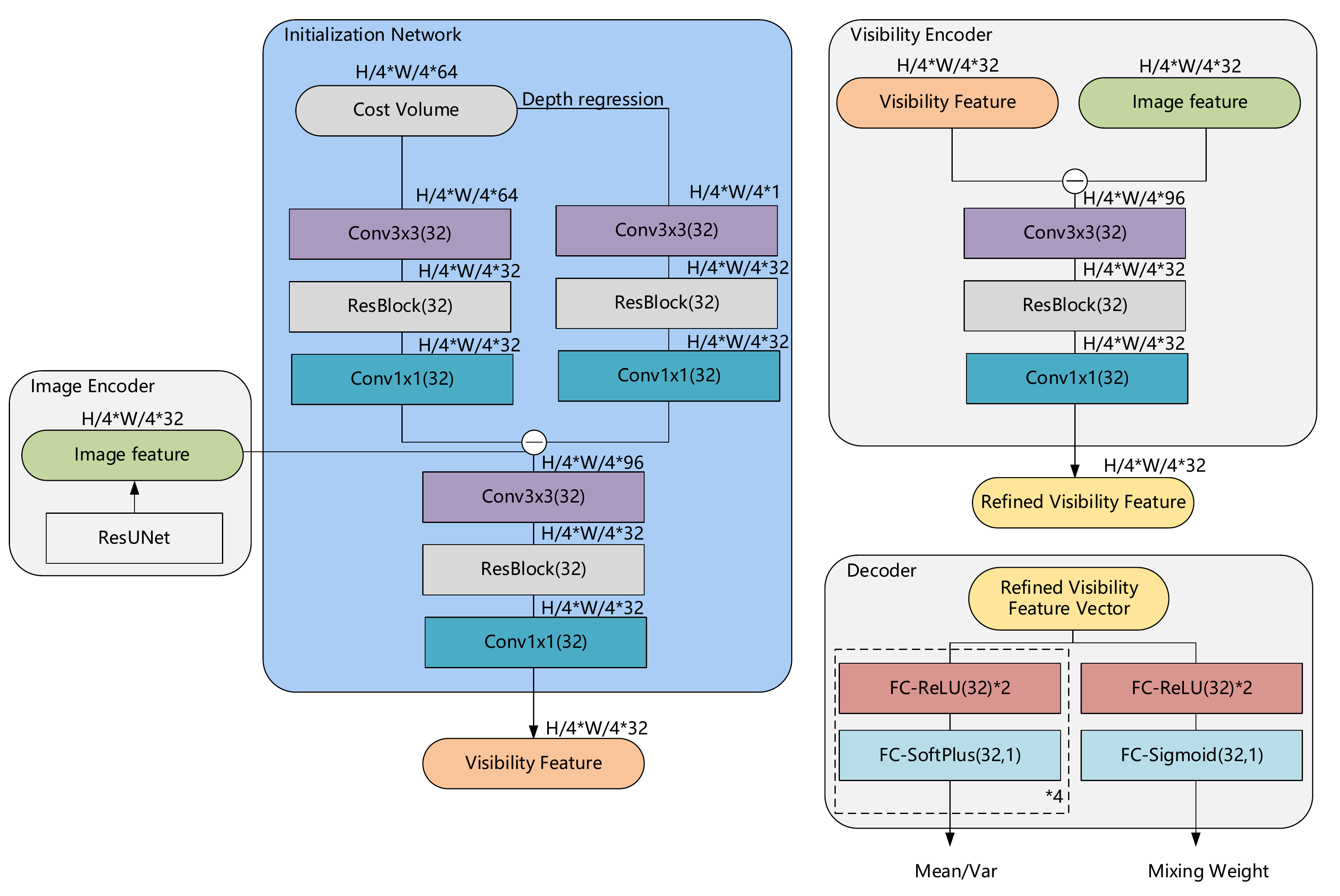}
    \caption{Architecture details of networks used in NeuRay.}
    \label{fig:arch1}
\end{figure*}

\begin{figure*}
    \centering
    \includegraphics[width=\textwidth]{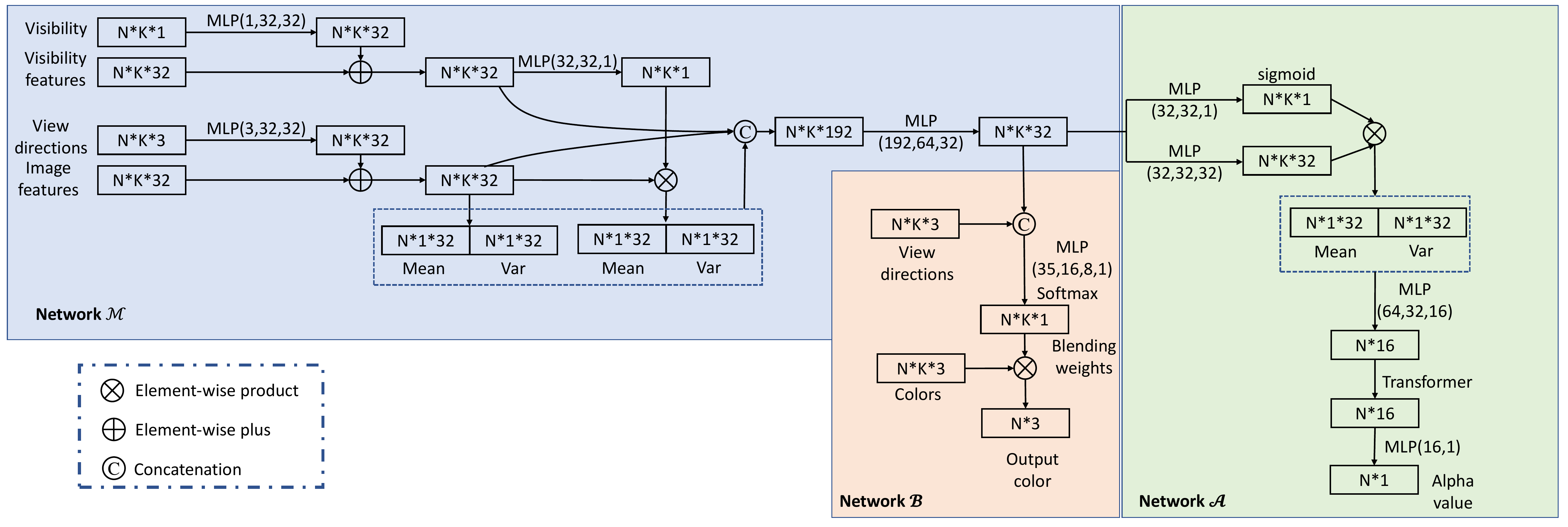}
    \caption{Architecture details of feature aggregation networks $(\mathcal{M,A,B})$. N is the number of sample points on a test ray and K is the number of working views.}
    \label{fig:arch2}
\end{figure*}

\textbf{Architecture}. In the cost volume construction, $N_s=3$ neighboring input views are used as source views and the height and width of the cost volume are $1/4$ of the height and width of the original image. The cost volume is constructed by a MVSNet~\cite{yao2018mvsnet} pretrained on the DTU training set. When training our renderer, we fixed the weight of this MVSNet due to GPU memory limitation but it is possible to train this part for better performance. The architecture is illustrated in Fig.~\ref{fig:arch1} and Fig.~\ref{fig:arch2}. The image encoder is a ResNet~\cite{he2016deep} with 13 residual blocks which outputs a feature map with 32 channels. All convolution layers use ReLU as activation function and all batch normalization layers are replaced by instance normalization layers. Compared to IBRNet~\cite{wang2021ibrnet}, our image encoder is more lightweight with less intermediate channel number and less residual blocks. The distribution decoder $\mathcal{F}$ contains 5 sub-networks, each of which consists of 3 fully-connected layers. These sub-networks are in charge of the prediction of $\mu_1$, $\mu_2$, $\sigma_1$, $\sigma_2$ and mixing weight $w_0$ respectively ($w_1=1-w_0$). All fully-connected layers in $\mathcal{F}$ use ReLU as the activation function except for the final layer which use SoftPlus or Sigmoid as the activation function. 

The architecture of feature aggregation networks follows design of IBRNet~\cite{wang2021ibrnet}, which is illustrated in Fig.~\ref{fig:arch2}. 
Based on the aggregated feature, the alpha value $\alpha_i$ is computed by
\begin{equation}
    \alpha_i = \mathcal{A} ({\bm{f}}_i),
\end{equation}
where $\mathcal{A}$ is an alpha network. The color $\bm{c}_i$ is computed by
\begin{equation}
    \bm{c}_i = \mathcal{B} ({\bm{f}}_i,\{\bm{f}_{i,j};\bm{r}_{i,j} - \bm{r};\bm{c}_{i,j};v_{i,j}\}),
\end{equation}
where $\mathcal{B}$ is a color blending network, $\bm{r}_{i,j}$ is the view direction from $j$-th input view to the point $\bm{p}_i$, and $\bm{c}_{i,j}$ is the color of this point projected on the input view. The alpha network $\mathcal{A}$ will learn to assign a large alpha value $\alpha_i$ to $\bm{p}_i$ when local image features $\{\bm{f}_{i,j}\}$ of all visible input views are consistent. Meanwhile, the color blending network $\mathcal{B}$ will learn to use the color $\bm{c}_{i,j}$ from visible views to produce the color $\bm{c}_i$ on $\bm{p}_i$.

\textbf{Depth loss}. In pretraining the generalization model on training scenes, we also force the mean $\mu_1$ of the first logistics distribution to be similar to the input depth by a depth loss $\ell_{depth}$
\begin{equation}
    \ell_{depth}=\sum \|\mu_1 - z_{in}\|^2,
\end{equation}
where the depth $z_{in}$ is computed from the cost volume or the input estimated depth maps.

\textbf{Training details}. We use Adam~\cite{kingma2014adam} as the optimizer with the default setting ($\beta_1=0.999$,$\beta_2=0.9$). On every training step, 512 rays are randomly sampled. The initial learning rate for the generalization model is 2e-4 which decays to its half on every 100k steps. We train the generalization model with 400k steps on the training set, which takes about 3 days on a single 2080Ti GPU. In the finetuning setting, the learning rate is initially set to 1e-4 and it also decays to its half after 100k training steps. Finetuning 200k steps on a NeRF synthetic dataset with resolution 800$\times$800 cost 20 hours but finetuning 200k steps on low resolution images with 400$\times$400 can be faster, which costs 8 hours.

\subsection{More comparison with baselines}

\begin{figure}
    \centering
    \setlength\tabcolsep{1.5pt}
    \begin{tabular}{ccc}
        GT & NeRF~\cite{mildenhall2020nerf} & NeuRay \\
         \includegraphics[width=0.32\linewidth]{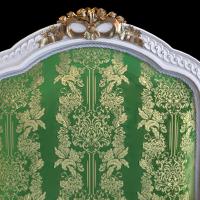} &  
         \includegraphics[width=0.32\linewidth]{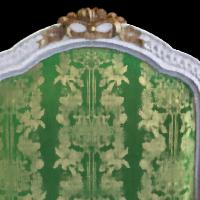} &  
         \includegraphics[width=0.32\linewidth]{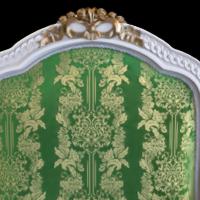} \\  
         \includegraphics[width=0.32\linewidth]{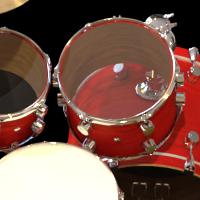} &  
         \includegraphics[width=0.32\linewidth]{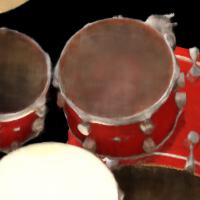} &  
         \includegraphics[width=0.32\linewidth]{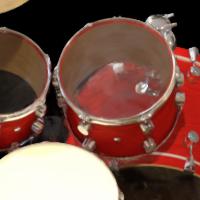} \\  
         \includegraphics[width=0.32\linewidth]{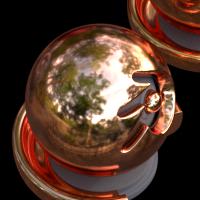} &  
         \includegraphics[width=0.32\linewidth]{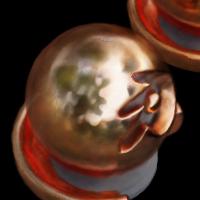} &  
         \includegraphics[width=0.32\linewidth]{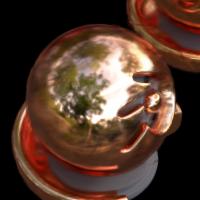} \\  
    \end{tabular}
    \caption{Qualitative comparison with NeRF~\cite{mildenhall2020nerf}.}
    \label{fig:nerf_compare}
\end{figure}

\begin{table}[]
    \centering
    \begin{tabular}{cccc}
    \toprule
    Dataset & Metrics & Cost volume & Depth maps \\
    \midrule
    \multirow{3}{*}{NeRF Syn.} 
    & PSNR & 28.29 & 28.92 \\
    & SSIM & 0.927 & 0.920 \\
    & LPIPS & 0.080 & 0.096 \\
    \midrule
    \multirow{3}{*}{DTU}
    & PSNR & 26.47 & 28.30 \\
    & SSIM & 0.875 & 0.907 \\
    & LPIPS & 0.158 & 0.130 \\
    \midrule
    \multirow{3}{*}{LLFF}
    & PSNR & 25.35 & 25.85 \\
    & SSIM & 0.818 & 0.832 \\
    & LPIPS & 0.198 & 0.190 \\
    \bottomrule
    \end{tabular}
    \caption{Results in the generalization setting using NeuRay initialized by estimated depth maps~\cite{schoenberger2016mvs} or constructed cost volumes~\cite{yao2018mvsnet} }
    \label{tab:depth}
\end{table}
\begin{table}[]
    \centering
    \begin{tabular}{cccc}
        \toprule
              & $N_w=8$ & $N_w=12$ & $N_w=16$ \\
         \midrule
         PSNR & 32.97   & 33.20    & 33.60    \\
         \bottomrule
    \end{tabular}
    \caption{PSNR on the Lego from the NeRF synthetic dataset with different working view numbers $N_w$.}
    \label{tab:more_vn}
\end{table}
\begin{table}[]
    \centering
    \begin{tabular}{cccc}
        \toprule
         Method & $N_w=8$ & $N_w=4$ & $N_w=2$ \\
         \midrule
         IBRNet~\cite{wang2021ibrnet} &
         26.87 & 25.41 & 21.72\\
         NeuRay &
         27.06 & 26.43 & 24.86\\
         \bottomrule
    \end{tabular}
    \caption{PSNR on the LLFF dataset with $N_w=2, 4, 8$ working views. The performance of IBRNet~\cite{wang2021ibrnet} drops with the decrease of working views while NeuRay still performs well with only 2 working views. All models are already finetuned on the scene for 200k steps.}
    \label{tab:wn}
\end{table}
\begin{table}[]
    \centering
    \begin{tabular}{ccccc}
        \toprule
        Setting & Method & $N=100$ & $N=75$ & $N=50$ \\
        \midrule
        \multirow{2}{*}{Gen} &
        NeuRay & 28.41 & 28.06 & 26.31 \\
        &
        IBRNet & 25.64 & 25.16 & 23.68 \\
        \midrule
        \multirow{2}{*}{Ft} &
        NeuRay & 32.97 & 32.71 & 31.40  \\
        &
        IBRNet & 30.37 & 28.57 & 25.68 \\
        
        \bottomrule
    \end{tabular}
    \caption{PSNR on the Lego with different input view numbers. With the decrease of view number, the performance decreases but NeuRay consistently outperforms IBRNet with different view numbers. ``Gen" means the generalization setting while ``Ft" means finetuning on the scene.}
    \label{tab:input_vn}
\end{table}
\begin{figure}
    \centering
    \begin{tabular}{ccc}
         $N=100$ & $N=75$ & $N=50$  \\
         \includegraphics[width=0.3\linewidth]{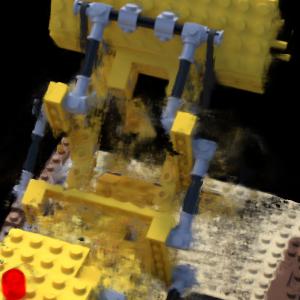} &
         \includegraphics[width=0.3\linewidth]{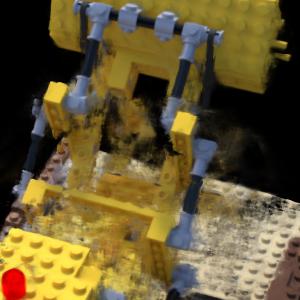} &
         \includegraphics[width=0.3\linewidth]{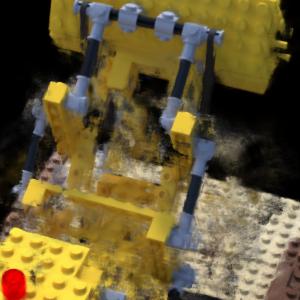} \\
         \includegraphics[width=0.3\linewidth]{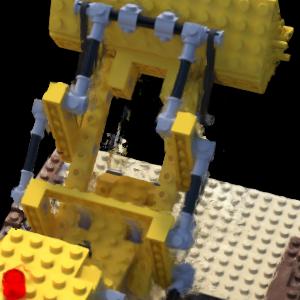} &
         \includegraphics[width=0.3\linewidth]{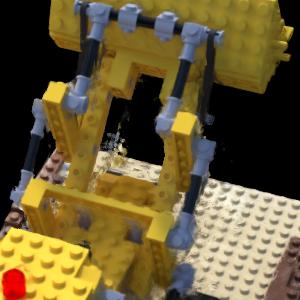} &
         \includegraphics[width=0.3\linewidth]{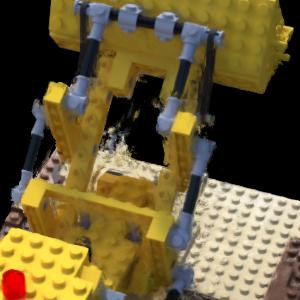} \\
    \end{tabular}
    \caption{Qualitative results in the generalization setting with different input view numbers. Row 1 shows the results of IBRNet~\cite{wang2021ibrnet} and Row 2 shows the results of NeuRay.}
    \label{fig:in_vn}
\end{figure}
\begin{figure}
    \centering
    \begin{tabular}{cc}
         Initial & Finetuned \\
         \includegraphics[width=0.45\linewidth]{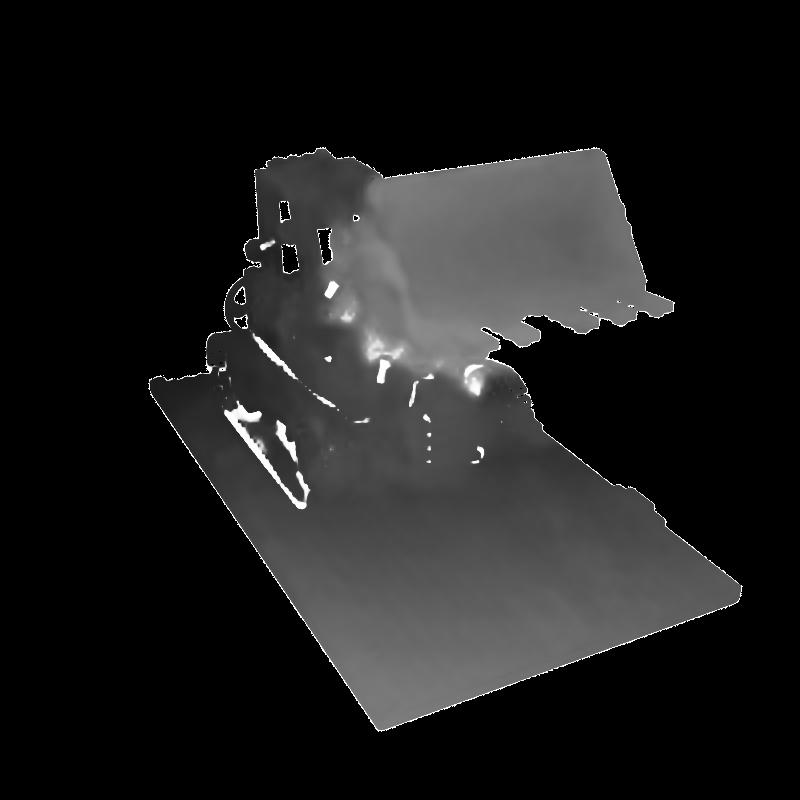}& 
         \includegraphics[width=0.45\linewidth]{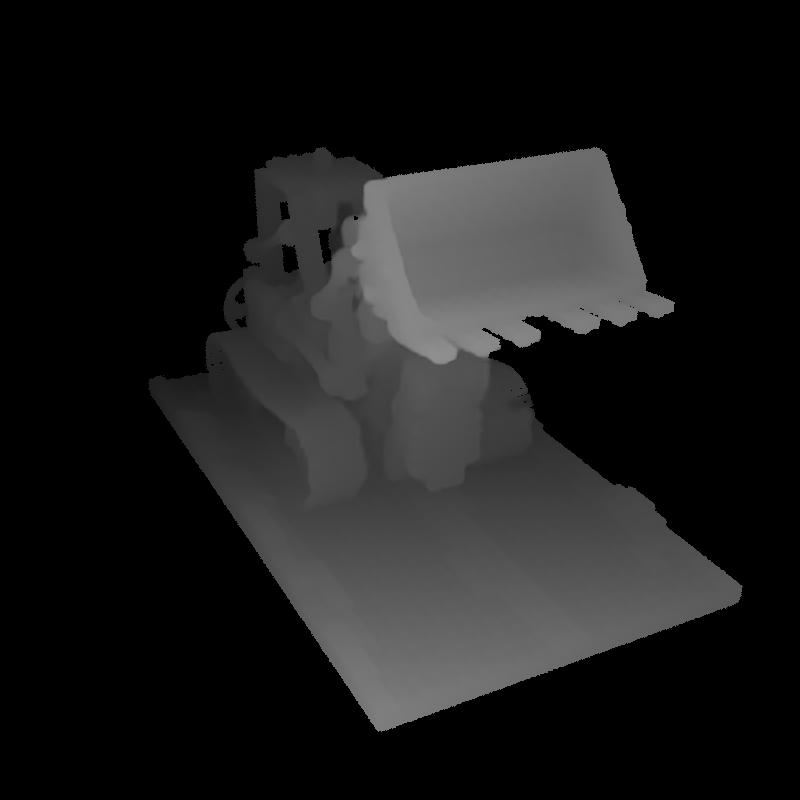}\\
         \includegraphics[width=0.45\linewidth]{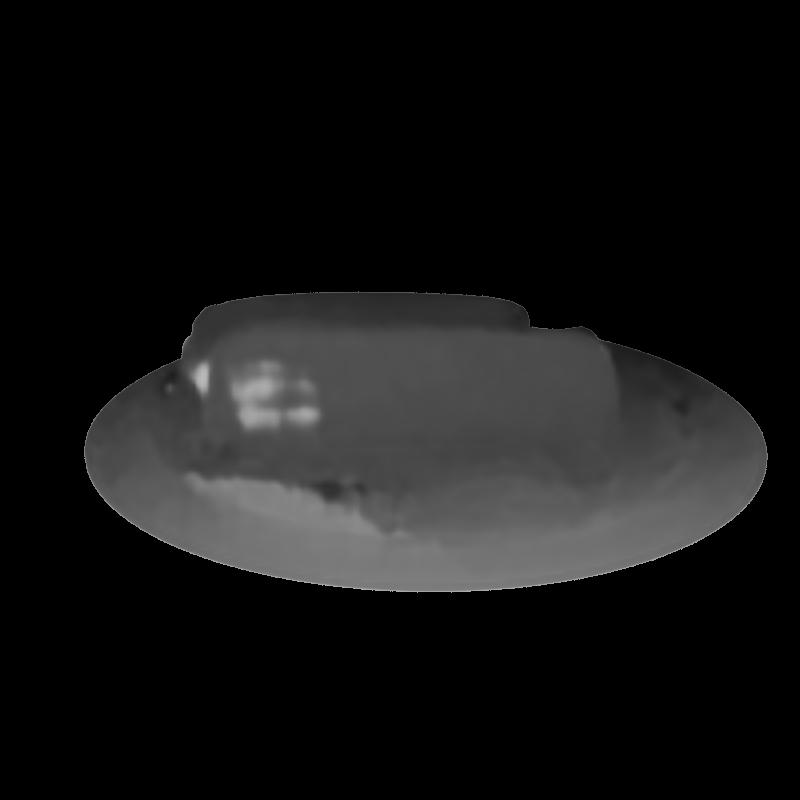}& 
         \includegraphics[width=0.45\linewidth]{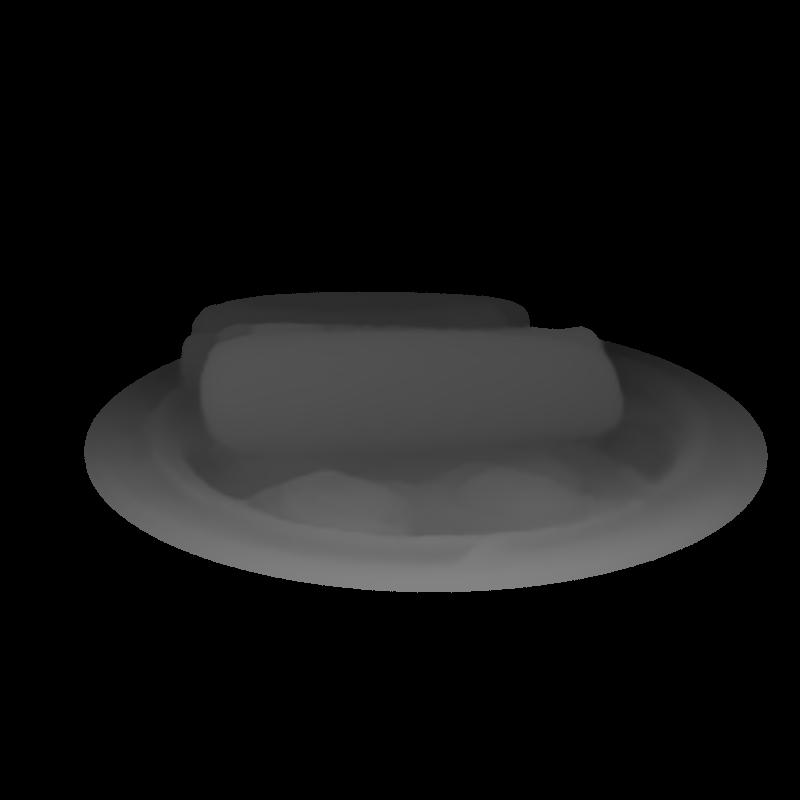}\\
    \end{tabular}
    \caption{Depth maps computed on the input views. Left are produced by initialized NeuRay and Right are produced by finetuned NeuRay.}
    \label{fig:depth}
\end{figure}

\textbf{Sparse working views on the LLFF dataset}. As discussed in the main paper, the reason that our method and IBRNet~\cite{wang2021ibrnet} perform very similar on the LLFF dataset is that the LLFF dataset contains very dense forward-facing views, which alleviates the feature inconsistency caused by occlusions. 
To demonstrate that, we decrease the number of working view $N_w$ to render the novel images. 
We show the qualitative results with 2 working views in Fig.~\ref{fig:work_num} and the PSNR on the LLFF dataset with different number of workings views in Table~\ref{tab:wn}. 
The results show that out method consistently outperforms IBRNet and the performance gap enlarges with the decrease of working view number. 
The reason is that with the decrease of working views, every 3D point is visible to input views so that the feature inconsistency caused by occlusions becomes more severe. 
Hence, because feature aggregation of IBRNet is not very robust to feature inconsistency, its performance degenerates. 
In contrast, with the constructed NeuRay, our method is occlusion-aware and thus performs better with sparse views. 
Meanwhile, we also find that undistortion by COLMAP~\cite{schoenberger2016mvs} is not perfect and noticeable distortions still remain in the LLFF dataset, which makes the scene-specific optimization of all methods struggles to improve.

\textbf{More qualitative comparison with NeRF}. In Fig.~\ref{fig:nerf_compare}, we show more comparison with NeRF~\cite{mildenhall2020nerf}, in which our method can recover details more clearly than NeRF with the same training steps on the given scene.

\subsection{Initialize from estimated depth maps}
Besides initialization from the cost volumes, NeuRay could also be initialized from estimated depth maps by patch match stereo in COLMAP~\cite{schoenberger2016mvs}.
In Table~\ref{tab:depth}, we show that initialization from estimated depth maps also produces good rendering results. Especially on the DTU dataset, COLMAP produces more accurate reconstruction than cost volumes so that the rendering quality is better.

\subsection{View consistency with more working views}
\textbf{Can more working views improve the rendering quality?}
We observe that simply using more working views on our trained model does not lead to obvious improvements, because the model is finetuned with 8 working views. However, finetuning our model with more working views indeed improves the rendering quality.
We finetune our model with different numbers of working views 8, 12, 16 on the Lego from the NeRF synthetic dataset. The results are shown in Table~\ref{tab:more_vn}. We can see that training with more working views improves the results. The reason is that adding more working views in training means using more views for the feature aggregation to reconstruct the scene geometry, which enforces the view consistency and improves the accuracy of estimated surfaces.

\subsection{Results with sparse input views}

To investigate how the performance degenerates as the decrease of the number of input views, we reduce the input views of the Lego of the NeRF synthetic dataset from 100 to 75 and 50 by the farthest point sampling on camera locations. The PSNR is reported in Table~\ref{tab:input_vn} and qualitative results are shown in Fig.~\ref{fig:in_vn}. The performance degrades reasonably as the input views become sparser but is consistently better than IBRNet.
% The qualitative results show that artifacts may appear when there are only 25 input views, which are too sparse to capture the detailed shape of the scene. 
% However, increasing the input view number to 50 already suffices for our method to produce images of satisfactory quality.

\subsection{Estimation of depth maps}
We can decode a depth map from NeuRay on a input view. Along every input ray, we evenly sample points and use the depth of the sample point with the largest $\tilde{h}_i$ as the depth of this input ray. In Fig.~\ref{fig:depth}, we show the depth maps from the initialized and the finetuned NeuRay representation. The results show that initialized NeuRay already produces a reasonable depth maps and optimizing NeuRay representation is able to enforce the consistency between views and rectify erroneous depth values. In light of this, finetuning NeuRay may also be regarded as a process of depth fusion or a consistency check for MVS algorithm, which takes coarse depth maps as input and outputs refined consistent depth maps. However, unlike commonly-used consistency check which simply discards erroneous depth, optimizing NeuRay is able to correct these erroneous depth.

\subsection{Additional ablation studies} 
We have conducted additional ablation studies on some of our network designs in Table~\ref{tab:ablation_more}.

\noindent\textbf{1) Adding $\ell_{consist}$ in the generalization training}. $\ell_{consist}$ is designed in scene-specific finetuning to memorize the geometry. Adding the loss in cross-scene generalization training produces similar results as the model without the loss.

\noindent\textbf{2) Optimize $\bm{G}'$ only or network parameters only}. Either leads to inferior results because only optimizing $\bm{G}'$ will disable refinement of feature aggregation while only optimizing networks will disable refinement of scene geometry in finetuning.

\noindent\textbf{3) Number of mixed distributions $N_l$}. 
$N_l$ limits the maximum number of semi-transparent surfaces to be represented on an input ray. However, we find that increasing $N_l$ from 2 to 3 does not bring significant improvements because there are few semi-transparent surfaces in most cases.

\noindent\textbf{4) Finetune $\bm{G}$ instead of $\bm{G}'$}.
Finetuning $\bm{G}'$ with visibility encoder produces better results than finetuning $\bm{G}$ (+0.5 PSNR). The reason is that the convolution layers in the visibility encoder associate nearby feature vectors  of trainable $\bm{G}'$ and these nearby pixels usually have similar visibility.

\begin{table}[]
    \centering
    \resizebox{\linewidth}{!}{
    \begin{tabular}{cccc}
        \toprule
        \multicolumn{2}{c}{Generalization} &
        \multicolumn{2}{c}{Finetuning}     \\
        Description         & PSNR & Description &  PSNR \\
        \midrule
        Default NeuRay               &28.41 & Default NeuRay             & 32.97 \\
        Add $\ell_{consist}$ &28.46 & Optimize vis./net. only & 32.25/31.90 \\
        $N_l=3$      &  28.44 & $N_l=3$ & 33.00 \\
        && Finetune $\bm{G}$ & 32.43 \\
        \bottomrule
    \end{tabular}
    }
    \caption{Additional ablation studies on the Lego from the NeRF Synthetic dataset, reported in PSNR.}
    \label{tab:ablation_more}
\end{table}

\subsection{Direct rendering from NeuRay}
\textbf{Can we render a novel view image from the NeuRay representation without constructing a radiance field?}

Yes, we can render an image directly from NeuRay, which is called the \textit{direct rendering} of NeuRay. Given a test ray $\bm{o}+z\bm{r}$ with sample points $\{\bm{p}_i\}$, we will compute the output color for this ray by $\hat{\bm{c}}=\sum_i \hat{\bm{c}}_i\hat{h}_i$. In Sec.~\ref{sec:compare}, we already show how to compute $\hat{h}$. The only remaining problem is how to compute the color $\hat{\bm{c}}_i$ of each sample point directly from NeuRay. To achieve this, we adopt a traditional Spherical Harmonics Fitting here.

We use a set of spherical harmonic functions as basis functions to fit a color function $R:\mathbb{S}^2\to \mathbb{R}^3$ on every point.
Specifically, we solve the following linear least squares problem
\begin{equation}
    \mathop{min}_{\bm{\theta}_i} \sum_j \tilde{h}_{i,j}(z_{i,j},z_{i,j}+l_i) \|R(\bm{r}_{i,j};\bm{\theta}_i) - \bm{c}_{i,j}\|^2+\bm{\theta}^\intercal_i\bm{\Lambda}\bm{\theta}_i,
    \label{eq:shf}
\end{equation}
where $\tilde{h}_{i,j}$ is the hitting probability of point $\bm{p}_i$ on the input ray emitted from the $j$-th input view, $R$ the color function, $\bm{\theta}_i$ the coefficients of spherical harmonic basis functions on this point, $\bm{r}_{i,j}$ and $\bm{c}_{i,j}$ are the viewing direction and color of the input ray emitted from the $j$-th view, respectively, $\bm{\theta}^\intercal_i\bm{\Lambda}\bm{\theta}_i$ a regularization term, and $\bm{\Lambda}$ a predefined diagonal matrix. This linear least squares problem has a closed-form solution. After finding the solution $\bm{\theta}_i$ to Eq.\eqref{eq:shf}, the color $\bm{c}_i$ is computed by
\begin{equation}
    \bm{c}_i=R(\bm{r};\bm{\theta}_i).
\end{equation}
Note that in Eq.~\eqref{eq:shf}, the color difference is weighted by hitting probabilities so that occluded input rays will not interfere the output colors. Meanwhile, since we fit a function $R$ on the sphere, NeuRay is able to represent anisotropic colors (or radiance) at a point. NeX~\cite{Wizadwongsa2021NeX} and PlencOctree~\cite{yu2021plenoctrees}, also use spherical harmonics functions as basis functions to represent colors. In implementation, we use spherical harmonics up to 3 degree and thus $\theta\in \mathbb{R}^{16}$ and the diagonal elements of the regularization matrix $\Lambda$ are $(0,0.001,0.005,0.01)$ for degree 0 to 3.

\begin{table}[]
    \centering
    \begin{tabular}{ccc}
    \toprule
         Method    & Fern  & Lego \\
         \midrule
         NeuRay-DR & 25.27 & 29.93 \\
         NeuRay-RF & 25.93 & 32.97 \\
         \bottomrule
    \end{tabular}
    \caption{PSNR of Direct Rendering (DR) of NeuRay and rendering from the constructed radiance field (RF) with NeuRay.}
    \label{tab:dr}
\end{table}
\begin{figure}
    \centering
    \begin{tabular}{cc}
         \includegraphics[width=0.45\linewidth]{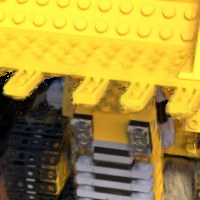} &
         \includegraphics[width=0.45\linewidth]{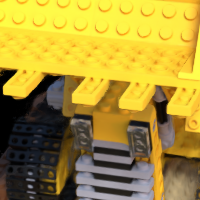} \\
         (a) & (b) 
    \end{tabular}
    \caption{Comparison between the direct rendering of NeuRay (a) and the rendering of the constructed radiance field (b). Direct rendering of NeuRay produces meaningful results but artifacts occur on the edges. Note the direct rendering is not supervised.}
    \label{fig:dr_compare}
\end{figure}
\begin{figure}
    \centering
    \begin{tabular}{cc}
         NeuRay & NeRF~\cite{mildenhall2020nerf} \\
         \includegraphics[width=0.4\linewidth]{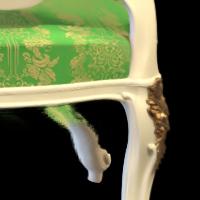} &
         \includegraphics[width=0.4\linewidth]{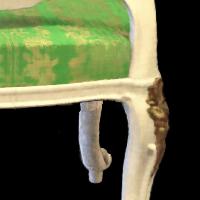} \\
    \end{tabular}
    \caption{As an image-based rendering method, NeuRay is unable to correctly render the region that is not visible to all working views, even though such regions are visible on other images. In comparison, as a global scene representation, NeRF~\cite{mildenhall2020nerf} can correctly render these regions.}
    \label{fig:limitation}
\end{figure}

\textbf{Quality of direct rendering}. To show the direct rendering of NeuRay can also produce reasonable renderings, we conduct experiments on the Lego and the Fern using the finetuned models. The qualitative results are shown in Fig.~\ref{fig:dr_compare} and the quantitative results are shown in Table~\ref{tab:dr}. The results demonstrate that though the direct rendering of NeuRay is not supervised by any rendering loss, it is still able to produce images with correct structures and details. Further supervising the direct rendering with a rendering loss may improve its rendering quality.

\subsection{Limitations}

As an image-based rendering method, our method needs to select a set of working views to render a novel view image. Hence, NeuRay is unable to render pixels that are invisible to all working views, though these pixels may be visible in other input views. We show an example in Fig.~\ref{fig:limitation}, in which the region is invisible to all working views. Meanwhile, our method is based on feature matching of input views to reconstruct the scene geometry. Such feature matching may struggle to find correct surfaces in textureless regions or cluttered complex regions like the Ficus. In this case, our method may not be able to keep the view consistency and correctly render, leading to rendering artifacts. Including more input views and working views or improving the feature matching may alleviate this problem. Another limitation is that speeding up rendering with NeuRay requires an accurate estimation of the surface locations so that such speeding up is only applicable to the finetuned model with accurate visibility. In the generalization setting, the predicted visibility is not accurate enough to maintain the rendering quality with very few sample points. 

\appendix

\end{document}